\documentclass{article}

\PassOptionsToPackage{numbers,compress}{natbib}
\usepackage[preprint]{neurips_2021}

\usepackage[utf8]{inputenc} % allow utf-8 input
\usepackage[T1]{fontenc}    % use 8-bit T1 fonts
\usepackage{hyperref}       % hyperlinks
\usepackage{url}            % simple URL typesetting
\usepackage{booktabs}       % professional-quality tables
\usepackage{amsfonts,amsmath}       % blackboard math symbols
\usepackage{nicefrac}       % compact symbols for 1/2, etc.
\usepackage{microtype}      % microtypography
\usepackage[table]{xcolor}
\usepackage{xcolor}         % colors
\usepackage{multirow}       % multiple rows in table
\usepackage{graphicx}
\usepackage{graphics}
\usepackage{tabularx}
\usepackage{makecell}
\usepackage{caption}
\usepackage{wrapfig}
\usepackage{enumitem}
\usepackage{subcaption}
\usepackage{algorithm}
\usepackage{algpseudocode}
\usepackage{amsfonts}       % blackboard math symbols
\usepackage{nicefrac}       % compact symbols for 1/2

\usepackage{color, colortbl}
\definecolor{Gray}{gray}{0.93}
\definecolor{orange}{rgb}{0.9,0.5,0}

\newcommand{\cocominivalboxmapms}{\textbf{58.7}}
\newcommand{\cocotestdevboxmapms}{\textbf{58.9}}
\newcommand{\adevalss}{\textbf{54.0}}
\newcommand{\adevalms}{\textbf{55.4}}
\title{Focal Self-attention for Local-Global Interactions in \\ Vision Transformers}

\author{%
  Jianwei Yang$^{1}$ \,\, Chunyuan Li$^{1}$ \,\, Pengchuan Zhang$^{1}$ \,\, Xiyang Dai$^{2}$ \,\, Bin Xiao$^{2}$ \\ 
  \textbf{Lu Yuan}$^{2}$ \,\, \textbf{Jianfeng Gao}$^{1}$ \\
  $^1$Microsoft Research at Redmond, $^2$Microsoft Cloud + AI\\
  \texttt{\{jianwyan,chunyl,penzhan,xidai,bixi,luyuan,jfgao\}@microsoft.com} \\
}

\begin{document}

\maketitle

\begin{abstract}
    Recently, Vision Transformer and its variants have shown great promise on various computer vision tasks. The ability of capturing short- and long-range visual dependencies through self-attention is  the key to success. But it also brings challenges due to quadratic computational overhead, especially for the high-resolution vision tasks (\textit{e.g.}, object detection). Many recent works have attempted to reduce the computational and memory cost \emph{and} improve performance by applying either coarse-grained global attentions or fine-grained local attentions. However, both approaches cripple the modeling power of the original self-attention mechanism of multi-layer Transformers, thus leading to sub-optimal solutions. In this paper, we present \emph{focal self-attention}, a new mechanism that incorporates both fine-grained local and coarse-grained global interactions. In this new mechanism, each token attends its closest surrounding tokens at fine granularity and the tokens far away at coarse granularity, and thus can capture both short- and long-range visual dependencies efficiently \textit{and} effectively. With focal self-attention, we propose a new variant of Vision Transformer models, called \emph{Focal Transformer}, which achieves superior performance over the state-of-the-art (SoTA) vision Transformers on a range of public image classification and object detection benchmarks. In particular, our Focal Transformer models with a moderate size of 51.1M and a larger size of 89.8M achieve 83.5\% and 83.8\% Top-1 accuracy, respectively, on ImageNet classification at $224 \times 224$. When employed as the backbones, Focal Transformers achieve consistent and substantial improvements over the current SoTA Swin Transformers~\cite{liu2021swin} across 6 different object detection methods. Our largest Focal Transformer yields \cocominivalboxmapms/\cocotestdevboxmapms~box mAPs and \textbf{50.9}/\textbf{51.3} mask mAPs on COCO mini-val/test-dev, and \adevalms~mIoU on ADE20K for semantic segmentation, creating new SoTA on three of the most challenging computer vision tasks.
    % Further studies show that both local and global visual dependencies contributes to the final performance.

\end{abstract}

\section{Introduction}
Nowadays, Transformer~\cite{vaswani2017attention} has become a prevalent model architecture in natural language processing (NLP)~\cite{devlin2019bert,brown2020language}. In the light of its success in NLP, there is an increasing effort on adapting it to computer vision (CV)~\cite{parmar2018image,ramachandran2019stand}. Since its promise firstly demonstrated in Vision Transformer (ViT)~\cite{dosovitskiy2020image}, we have witnessed a flourish of full-Transformer models for image classification~\cite{touvron2020training,wang2021pyramid,wu2021cvt,liu2021swin,zhang2021multi,vaswani2021scaling}, object detection~\cite{carion2020end,zhu2020deformable,zheng2020end,dai2020up} and semantic segmentation~\cite{wang2020max,wang2020end}. Beyond these static image tasks, it has also been applied on various temporal understanding tasks, such as action recognition~\cite{li2021trear,zhao2021tuber,chang2021augmented}, object tracking~\cite{chen2021Transformer,wang2021Transformer}, scene flow estimation~\cite{li2021sctn}. 
% The competitive or superior performance on this brand spectrum of vision tasks imply a potential model shift from the widely used Convolutional Neural Networks (CNN)

\begin{figure}[t!]
\centering
    \includegraphics[height=0.28\textwidth]{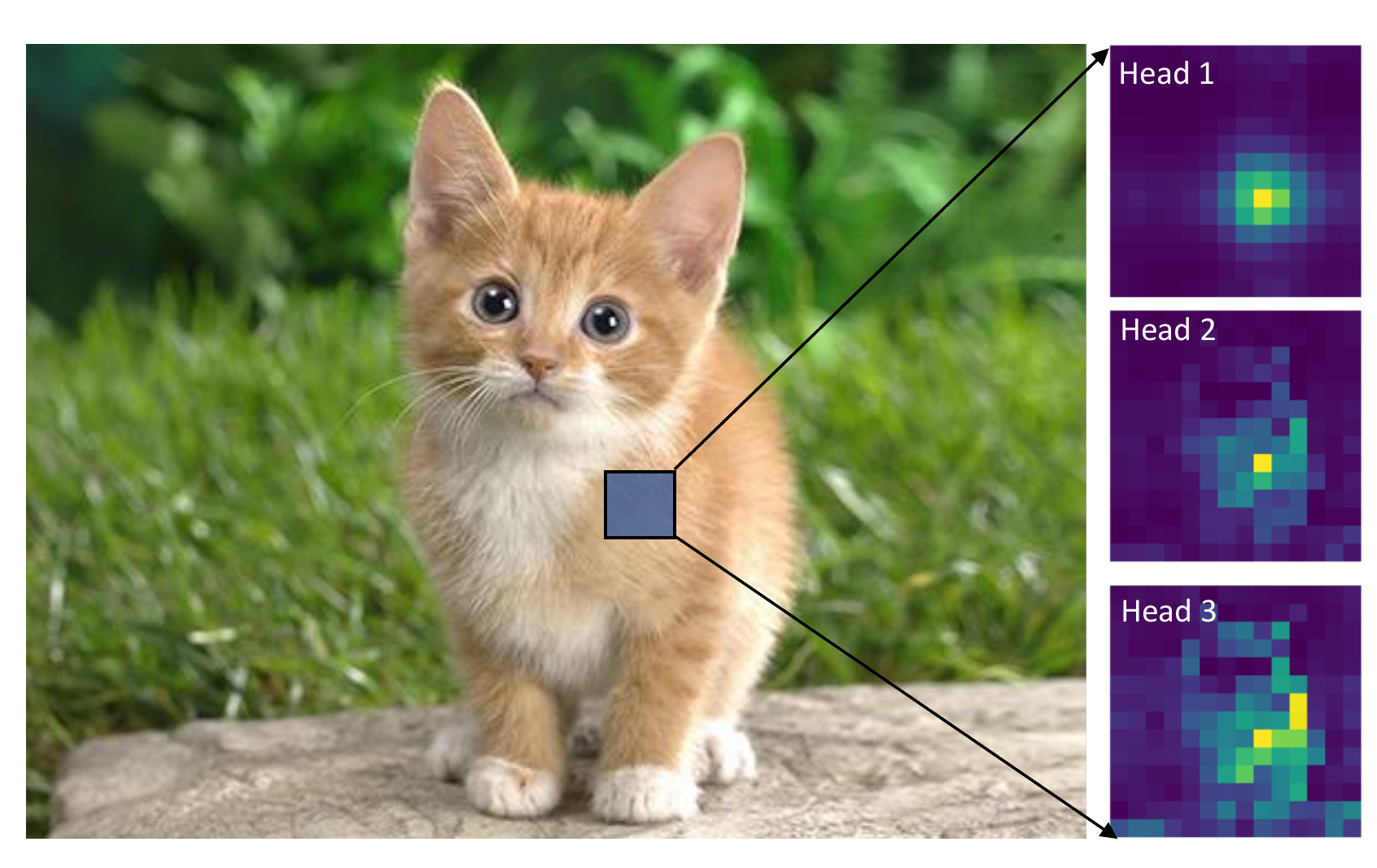}
    \quad
    \includegraphics[height=0.28\textwidth]{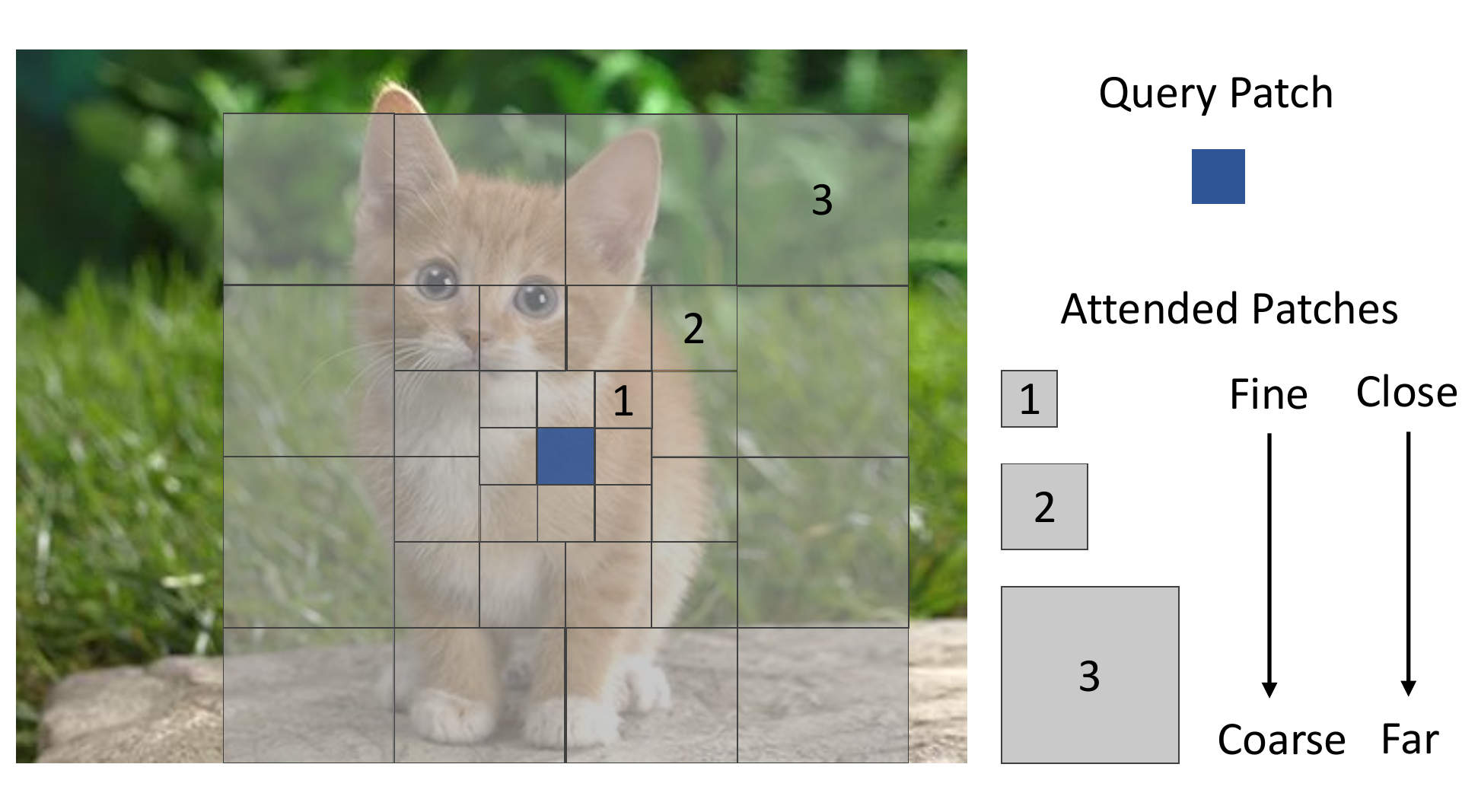}
    \vspace{-3mm}
  \caption{Left: Visualization of the attention maps of the three heads at the given query patch (blue) in the first layer of the DeiT-Tiny model~\cite{touvron2020training}. Right: An illustrative depiction of focal self-attention mechanism. Three granularity levels are used to compose the attention region for the blue query.}
  \label{fig:teaser_fig}
  \vspace{-4mm}
\end{figure}

In Transformers, self-attention is the key component making it unique from the widely used convolutional neural networks (CNNs)~\cite{lecun1995convolutional}. At each Transformer layer, it enables the global content-dependent interactions among different image regions for modeling both short- and long-range dependencies. Through the visualization of full self-attentions\footnote{DeiT-Tiny model, checkpoint downloaded from \url{https://github.com/facebookresearch/deit}.}, we indeed observe that it learns to attend local surroundings (like CNNs) and the global contexts at the same time (See the left side of Fig.~\ref{fig:teaser_fig}). Nevertheless, when it comes to high-resolution images for dense predictions such as object detection or segmentation, a global and fine-grained self-attention becomes non-trivial due to the quadratic computational cost with respect to the number of grids in feature maps. 
Recent works alternatively exploited either a coarse-grained global self-attention~\cite{wang2021pyramid,wu2021cvt} or a fine-grained local self-attention~\cite{liu2021swin,zhang2021multi,vaswani2021scaling} to reduce the computational burden. However, both approaches  
cripple the power of the original full self-attention 
%after sacrificing one of its two merits, 
\textit{i.e.}, the ability to simultaneously model short- and long-range visual dependencies, as demonstrated on the left side of Fig.~\ref{fig:teaser_fig}.

In this paper, we present a new self-attention mechanism to capture both local and global interactions in Transformer layers for high-resolution inputs. Considering that the visual dependencies between regions nearby are usually stronger than those far away, we perform the fine-grained self-attention only in local regions while the coarse-grained attentions globally. 
As depicted in the right side of Fig.~\ref{fig:teaser_fig}, a query token in the feature map attends its closest surroundings at the finest granularity as itself. However, when it goes to farther regions, it attends to summarized tokens to capture coarse-grained visual dependencies. 
The further away the regions are from the query, the coarser the granularity is 
%information coarsely. 
%By this way, we can further cover the whole feature map using coarser-grained at even farther distance. 
As a result, it can effectively cover the whole high-resolution feature maps while introducing much less number of tokens in the self-attention computation than that in the full self-attention mechanism. As a result, it has the ability to capture both short- and long-range visual dependencies efficiently.
We call this new mechanism {\it focal self-attention}, as each token attends others in a focal manner. 
% Clearly, this focal self-attention has the ability to capture both short- and long-range visual dependencies while introducing moderate computational overhead after coarsening the tokens in the feature map.
% 
Based on the proposed focal self-attention, a series of Focal Transformer models are developed, by 1) exploiting a multi-scale architecture to maintain a reasonable computational cost for high-resolution images~\cite{wang2021pyramid,wu2021cvt,liu2021swin,zhang2021multi}, and 2) splitting the feature map into multiple windows in which tokens share the same surroundings, instead of performing focal self-attention for each token~\cite{vaswani2021scaling,zhang2021multi,liu2021swin}. 

We validate the effectiveness of the proposed focal self-attention via a comprehensive empirical study on  image classification, object detection and segmentation. 
Results show that our Focal Transformers with similar model sizes and complexities consistently outperform the SoTA Vision Transformer models across various settings. Notably, our small Focal Transformer model with 51.1M parameters can achieve 83.5\% top-1 accuracy on ImageNet-1K, and the base model with 89.8M parameters obtains 83.8\% top-1 accuracy. When transferred to object detection, our Focal Transformers consistently outperform the SoTA Swin Transformers~\cite{liu2021swin} for six different object detection methods. Our largest Focal Transformer model achieves \cocotestdevboxmapms~box mAP and \textbf{51.3} mask mAP on COCO test-dev for object detection and instance segmentation, respectively, and \adevalms~mIoU on ADE20K for semantic segmentation. These results demonstrate that the focal self-attention is highly effective in  modeling the local-global interactions in Vision Transformers.

\section{Method}
\vspace{-2mm}
\begin{figure}[t!]
\centering
    \includegraphics[width=1.0\textwidth]{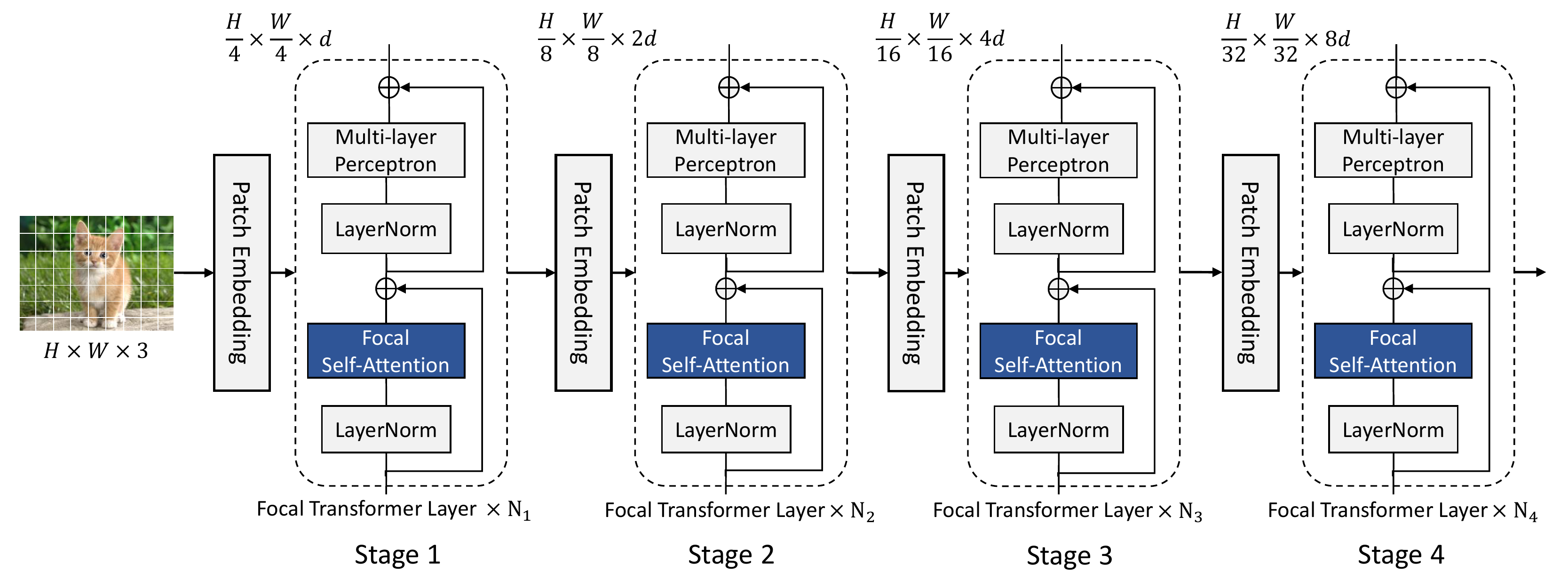}
  \caption{Model architecture for our Focal Transformers. As highlighted in light blue boxes, our main innovation is the proposed focal self-attention mechanism in each Transformer layer.}
  \label{fig:model_architecture}
\end{figure}
% \begin{table}[!ht]
% \centering
%     \begin{tabular}{lccccc}
%     \toprule
%     Method & Image Size & \#Params. & GFLOPs & Thr. (images/s) & Top-1 Acc. \\
%     \midrule
%     ViT~\cite{} & \\
%     CPVT~\cite{} & \\
%     T2T~\cite{} & \\
%     PVT~\cite{} & \\
%     CvT~\cite{} & \\
%     \bottomrule
%     \end{tabular}
%     \caption{Model architecture for focal vision Transformer.}
%     \label{tab:self_attention}
% \end{table}
\subsection{Model architecture}

To accommodate the high-resolution vision tasks, our model architecture shares a similar multi-scale design with~\cite{wang2021pyramid,zhang2021multi,liu2021swin}, which allows us to obtain high-resolution feature maps at earlier stages. As shown in Fig.~\ref{fig:model_architecture}, an image $I \in \mathcal{R}^{H \times W \times 3}$ is first partitioned into patches of size $4\times 4$, resulting in $\frac{H}{4} \times \frac{W}{4}$ visual tokens with dimension $4 \times 4 \times 3$. Then, we use a patch embedding layer which consists of a convolutional layer with filter size and stride both equal to 4, to project these patches into hidden features with dimension $d$. Given this spatial feature map, we then pass it to four stages of focal Transformer blocks. At each stage $i \in \{1,2,3,4\}$, the focal Transformer block consists of $N_i$ focal Transformer layers. After each stage, we use another patch embedding layer to reduce the spatial size of feature map by factor 2, while the feature dimension is increased by 2. For image classification tasks, we take the average of the output from last stage and send it to a classification layer. For object detection, the feature maps from last 3 or all 4 stages are fed to the detector head, depending on the particular detection method we use. The model capacity can be customized by varying the input feature dimension $d$ and the number of focal Transformer layers at each stage $\{N_1, N_2, N_3, N_4\}$.

Standard self-attention can capture both short- and long-range interactions at fine-grain, but it suffers from high computational cost when it performs the attention on high-resolution feature maps as noted in~\cite{zhang2021multi}. 
% Suppose we have an input feature map $x\in \mathcal{R}^{M \times N \times d}$, where $M\times N$ is the spatial dimension. A global self-attention will bring the complexity of $\mathcal{O}((M\times N)^2d)$. This burden is particularly heavy when we have high resolution feature map. 
Take stage 1 in Fig.~\ref{fig:model_architecture} as the example. For the feature map of size $\frac{H}{4} \times \frac{W}{4} \times d$, the complexity of self-attention is $\mathcal{O}((\frac{H}{4}\times \frac{W}{4})^2d)$, resulting in an explosion of time and memory cost considering $\min(H, W)$ is 800 or even larger for object detection. In the next, we describe how we address this with the proposed focal self-attention.

\subsection{Focal self-attention}

\begin{wrapfigure}{R}{0.4\textwidth}
\vspace{-5mm}
\begin{minipage}{0.4\textwidth}
\scriptsize
\centering
\hspace{-3mm}
\includegraphics[height=2.6cm]{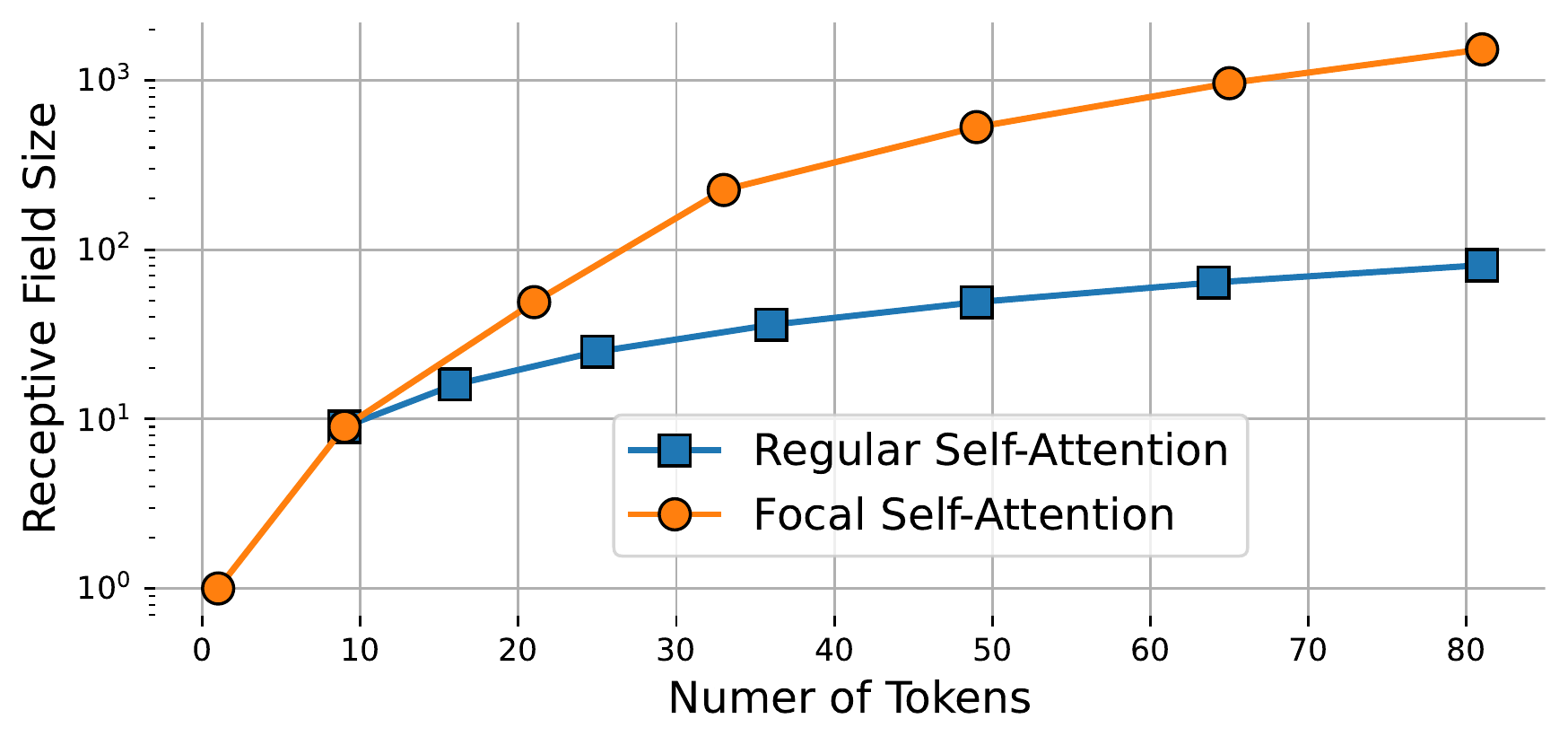}
\vspace{-1mm}
\caption{The size of receptive field (y-axis) with the increase of used tokens (x-axis) for standard and our focal self-attention. For focal self-attention, we assume increasing the window granularity by factor 2 gradually but no more than 8. Note that the y-axis is logarithmic.} 
\label{fig:receiptive_size_with_tokens}
\vspace{-3mm}
\end{minipage}
\end{wrapfigure}

In this paper, we propose focal self-attention to make Transformer layers scalable to high-resolution inputs. Instead of attending all tokens at fine-grain, we propose to attend the fine-grain tokens only locally, but the summarized ones globally. As such, it can cover as many regions as standard self-attention but with much less cost. In Fig.~\ref{fig:receiptive_size_with_tokens}, we show the area of receptive field for standard self-attention and our focal self-attention when we gradually add more attended tokens. For a query position, when we use gradually coarser-grain for its far surroundings, focal self-attention can have significantly larger receptive fields at the cost of attending the same number of visual tokens than the baseline. 

Our focal mechanism enables long-range self-attention with much less time and memory cost, because it attends a much smaller number of surrounding (summarized) tokens. In practice, however, extracting the surrounding tokens for each query position suffers from high time and memory cost since we need to duplicate each token for all queries that can get access to it. This practical issue has been noted by a number of previous works~\cite{vaswani2021scaling,zhang2021multi,liu2021swin} and the common solution is to partition the input feature map into windows. Inspired by them, we resort to perform focal self-attention at the window level. Given a feature map of $x \in \mathcal{R}^{M \times N \times d}$ with spatial size $M\times N$, we first partition it into a grid of windows with size $s_p \times s_p$. Then, we find the surroundings for each window rather than individual tokens. In the following, we elaborate the window-wise focal self-attention.

\subsubsection{Window-wise attention}

\begin{figure}[t]
\centering
    \includegraphics[width=0.95\textwidth]{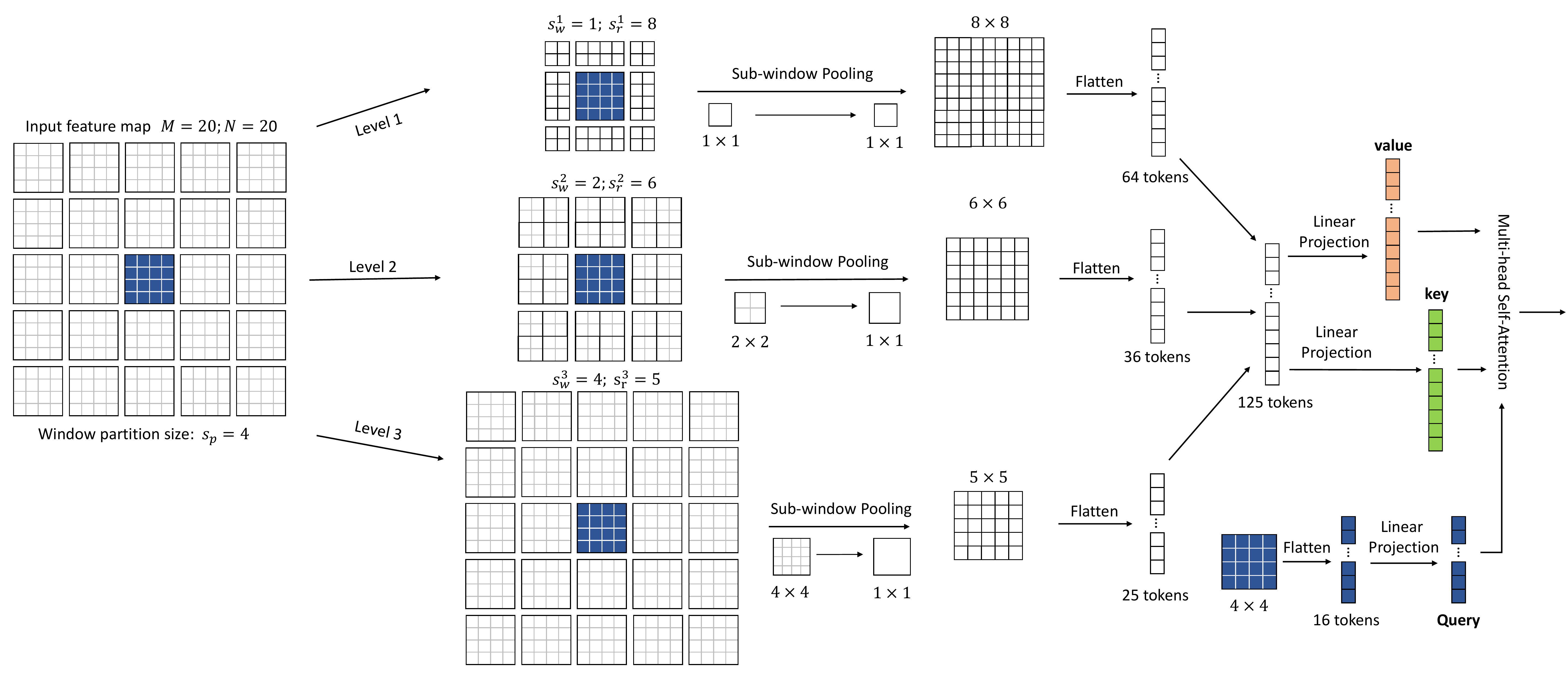}
  \caption{An illustration of our focal self-attention at window level. Each of the finest square cell represents a visual token either from the original feature map or the squeezed ones. Suppose we have an input feature map of size $20 \times 20$. We first partition it into $5 \times 5$ windows of size $4 \times 4$. Take the $4 \times 4$ blue window in the middle as the query, we extract its surroundings tokens at multiple granularity levels as its keys and values. For the first level, we extract the $8 \times 8$ tokens which are closest to the blue window at the finest grain. Then at the second level, we expand the attention region and pool the surrounding $2 \times 2$ sub-windows, which results in $6 \times 6$ pooled tokens. At the third level, we attend even larger region covering the whole feature map and pool $4\times 4$ sub-windows. Finally, these three levels of tokens are concatenated to compute the keys and values for the $4\times4 =16$ tokens (queries) in the blue window.}
  \label{fig:window_focal_attention}
\end{figure}

An illustration of the proposed window-wise focal self-attention is shown in Fig.~\ref{fig:window_focal_attention}. We first define three terms for clarity:
\begin{itemize}[noitemsep,topsep=0pt,leftmargin=*]
    \item \textbf{Focal levels} $L$ -- the number of granularity levels we extract the tokens for our focal self-attention. In Fig.~\ref{fig:teaser_fig}, we show 3 focal levels in total for example.
    \item \textbf{Focal window size} $s_w^l$ -- the size of sub-window on which we get the summarized tokens at level $l \in \{1,...,L\}$, which are 1, 2 and 4 for the three levels in Fig.~\ref{fig:teaser_fig}.
    \item \textbf{Focal region size} $s_r^l$ -- the number of sub-windows horizontally and vertically in attended regions at level $l$, and they are 3, 4 and 4 from level 1 to 3 in Fig.~\ref{fig:teaser_fig}.
\end{itemize}

% Though our focal self-attention illustrated in Fig.~\ref{fig:teaser_fig} consistently doubles the size of attended region and sub-window after each level, it is flexible to set them independently for each level. Another practical 
With the above three terms $\{L, s_w, s_r\}$, we can specify our focal self-attention module, proceeded in two main steps:

\textbf{Sub-window pooling}. Assume the input feature map $x \in \mathcal{R}^{M \times N \times d}$, where $M\times N$ are the spatial dimension and $d$ is the feature dimension. We perform sub-window pooling for all $L$ levels. For the focal level $l$, we first split the input feature map $x$ into a grid of sub-windows with size $s_w^l \times s_w^l$. Then we use a simple linear layer $f_p^l$ to pool the sub-windows spatially by:
\begin{equation}
\small
    x^l = f^l_p(\hat{x}) \in \mathcal{R}^{\frac{M}{s_w^l} \times \frac{N}{s_w^l} \times d}, \quad \hat{x}=\text{Reshape}(x)\in \mathcal{R}^{(\frac{M}{s_w^l} \times \frac{N}{s_w^l} \times d) \times (s_w^l \times s_w^l)},  
    \label{eq:pool}
\end{equation}

% Since $s^1_w=1$ at the first level, we skip the sub-window pooling, and directly copy it from $x$. 
The pooled feature maps $\{x^l\}_1^L$ at different levels $l$ provide rich information at both fine-grain and coarse-grain. Since we set $s_w^l=1$ for the first focal level which has the same granularity as the input feature map, there is no need to perform any sub-window pooling. Considering the focal window size is usually very small (7 maximally in our settings), the number of extra parameters introduced by these sub-window pooling are fairly negligible.

\textbf{Attention computation}. Once we obtain the pooled feature maps $\{x^l\}_1^{L}$ at all $L$ levels, we compute the query at the first level and key and value for all levels using three linear projection layers $f_q$, $f_k$ and $f_v$:
\begin{equation}
\small
    Q = f_q(x^1), \quad K = \{K^l\}_1^L =  f_k(\{x^1,...,x^L\}), \quad V = \{V^l\}_1^L = f_v(\{x^1,...,x^L\})
\end{equation}

To perform focal self-attention, we need to first extract the surrounding tokens for each query token in the feature map. As we mentioned earlier, tokens inside a window partition $s_p \times s_p$ share the same set of surroundings. For the queries inside the $i$-th window $Q_i \in \mathcal{R}^{s_p \times s_p \times d}$, we extract the $s_r^l \times s_r^l$ keys and values from $K^l$ and $V^l$ around the window where the query lies in, and then gather the keys and values from all $L$ to obtain $K_i = \{K_i^1, ..., K_i^L\} \in \mathcal{R}^{s \times d}$ and $V_i = \{V_i^1, ..., V_i^L\} \in \mathcal{R}^{s \times d}$, where $s$ is the sum of focal region from all levels, \textit{i.e.,}, $s=\sum_{l=1}^{L} (s_r^l)^2$. Note that a strict version of focal self-attention following Fig.~\ref{fig:teaser_fig} requires to exclude the overlapped regions across different levels. In our model, we intentionally keep them in order to capture the pyramid information for the overlapped regions. Finally, we follow~\cite{liu2021swin} to include a relative position bias and compute the focal self-attention for $Q_i$ by:
\begin{equation}
\small
    \text{Attention}(Q_i, K_i, V_i) = \text{Softmax}(\frac{Q_i K_i^T}{\sqrt{d}} + B)V_i,
    \label{eq:fsa}
\end{equation}
where $B = \{B^l\}_1^L$ is the learnable relative position bias. It consists of $L$ subsets for $L$ focal levels. Similar to~\cite{liu2021swin}, for the first level, we parameterize it to $B^1 \in \mathcal{R}^{(2s_p-1) \times (2s_p-1)}$, considering the horizontal and vertical position range are both in $[-s_p+1, s_p-1]$. For the other focal levels, considering they have different granularity to the queries, we treat all the queries inside a window equally and use $B^l \in \mathcal{R}^{s_r^l \times s_r^l}$ to represent the relative position bias between the query window and each of $s_r^l \times s_r^l$ pooled tokens. Since the focal self-attention for each window is independent of others, we can compute Eq.~\eqref{eq:fsa} in parallel. Once we complete it for the whole input feature map, we send it to the MLP block for proceeding computation as usual.

\subsubsection{Complexity analysis}

We analyze the computational complexity for the two main steps discussed above. For the input feature map $x \in \mathcal{R}^{M \times N \times d}$, we have $\frac{M}{s_w^l} \times \frac{N}{s_w^l}$ sub-windows at focal level l. For each sub-window, the pooling operation in Eq.\ref{eq:pool} has the complexity of $\mathcal{O}((s_w^l)^2d)$. Aggregating all sub-windows brings us $\mathcal{O}((MN)d)$. Then for all focal levels, we have the complexity of $\mathcal{O}(L(MN)d)$ in total, which is independent of the sub-window size at each focal level. Regarding the attention computation in Eq.~\ref{eq:fsa}, the computational cost for a query window $s_p\times s_p$ is $\mathcal{O}((s_p)^2 \sum_l (s_r^l)^2d)$, and $\mathcal{O}(\sum_l (s_r^l)^2(MN)d)$ for the whole input feature map. To sum up, the overall computational cost for our focal self-attention becomes $\mathcal{O}((L+\sum_l (s_r^l)^2)(MN)d)$. In an extreme case, one can set $s_r^L=2\max(M,N)/s_w^L$ to ensure global receptive field for all queries (including both corner and middle queries) in this layer. 
% At the extreme case, we can make $s_w^L=\max(M, N)$ which reduces to a global pooling over the whole feature map. 
% In a normal case that  $s_w^L=7$, we need roughly $2^2/7^2=8\%$ computational cost of the standard fully global self-attention to ensure global receptive field for all layers.

% \subsubsection{Relation to Previous Works}

% We particularly discuss the relation between focal self-attention with several most relevant prior arts:

% \begin{itemize}[noitemsep,topsep=0pt,leftmargin=*]
%     \item HaloNet~\cite{vaswani2021scaling} and ViL~\cite{zhang2021multi}. Both works exploited a block/window local attention, which first split input feature maps into blocks and then expands the block to its neighbor tokens. In HaloeNet, 
%     \item Swin Transformer~\cite{liu2021swin}. When we remove all coarse-grain attentions to merely keep the fine-grained 
%     \item PVT~\cite{wang2021pyramid} is at the other extreme of the spectrum. In their method, a spatial-reduction attention layer is used to reduce the number of keys and values by a per-determined factor. This operation is similar to our coarse-grain global attention when we increase the focal region size to cover the whole feature map. However, 
% \end{itemize}

% \subsubsection{Implementation Details}

% A practical challenge behind our focal self-attention is that we need to extract the surroundings from different levels for each window partition in the feature map. 

\subsection{Model configuration}

% $\begin{array}[t]#1{@{}>{$}c<{$}@{}}#2
%   #3\\#4
%   \end{array}$
  
\begin{table}[t!]
\setlength{\tabcolsep}{2.1pt}
\footnotesize
    \centering
    \resizebox{0.94\linewidth}{!}{
    \begin{tabular}{l|c|c|c|c|c}
    \toprule
            &  Output Size & Layer Name & Focal-Tiny & Focal-Small & Focal-Base \\
            \midrule
    \multirow{4}{*}{stage 1}  & $56\times56$    & Patch Embedding  &  $p_1=4;c_1=96$ & $p_1=4;c_1=96$ & $p_1=4;c_1=128$ \\
    \cmidrule{2-6}
                              & $56 \times 56$  &  \makecell{Transformer \\ Block} & \multicolumn{1}{c}{$\left[\begin{array}{c}
                                   s_{w,r}^{0}=\{1,13\} \\
                                   s_{w,r}^{1}=\{7,7\} 
                              \end{array}\right] \times 2$} &
\multicolumn{1}{c}{$\left[\begin{array}{c}
                                   s_{w,r}^{0}=\{1,13\} \\
                                   s_{w,r}^{1}=\{7,7\} 
                              \end{array}\right] \times 2$} &
\multicolumn{1}{c}{$\left[\begin{array}{c}
                                   s_{w,r}^{0}=\{1,13\} \\
                                   s_{w,r}^{1}=\{7,7\} 
                              \end{array}\right] \times 2$}
                              \\
        \midrule
    \multirow{4}{*}{stage 2} & $ 28\times 28$ & Patch Embedding & $p_2=2;c_2=192$ & $p_2=2;c_2=192$ & $p_2=2;c_2=256$ \\
    \cmidrule{2-6}
            & $ 28\times 28$ & \makecell{Transformer \\ Block} & \multicolumn{1}{c}{$\left[\begin{array}{c}
                                   s_{w,r}^{0}=\{1,13\} \\
                                   s_{w,r}^{1}=\{7,5\} 
                              \end{array}\right] \times 2$} & 
\multicolumn{1}{c}{$\left[\begin{array}{c}
                                   s_{w,r}^{0}=\{1,13\} \\
                                   s_{w,r}^{1}=\{7,5\} 
                              \end{array}\right] \times 2$} &
\multicolumn{1}{c}{$\left[\begin{array}{c}
                                   s_{w,r}^{0}=\{1,13\} \\
                                   s_{w,r}^{1}=\{7,5\} 
                              \end{array}\right] \times 2$}                              \\
        \midrule
    \multirow{4}{*}{stage 3} & $ 14\times 14$ & Patch Embedding & $p_3=2;c_3=384$ & $p_3=2;c_3=384$ & $p_3=2;c_3=512$\\
        \cmidrule{2-6}
            & $ 14\times 14$ & \makecell{Transformer \\ Block} & 
\multicolumn{1}{c}{$\left[\begin{array}{c}
                                   s_{w,r}^{0}=\{1,13\} \\
                                   s_{w,r}^{1}=\{7,3\} 
                              \end{array}\right] \times 6$} & \multicolumn{1}{c}{$\left[\begin{array}{c}
                                   s_{w,r}^{0}=\{1,13\} \\
                                   s_{w,r}^{1}=\{7,3\} 
                              \end{array}\right] \times 18$} & 
                 \multicolumn{1}{c}{$\left[\begin{array}{c}
                                   s_{w,r}^{0}=\{1,13\} \\
                                   s_{w,r}^{1}=\{7,3\} 
                              \end{array}\right] \times 18$}\\
        \midrule
    \multirow{4}{*}{stage 4} & $ 7\times 7$ & Patch Embedding & $p_4=2;c_4=768$ & $p_4=2;c_4=768$ & $p_4=2;c_4=1024$\\
    \cmidrule{2-6}
            & $ 7\times 7$ & \makecell{Transformer \\ Block} & \multicolumn{1}{c}{$\left[\begin{array}{c}
                                   s_{w,r}^{0}=\{1,7\} \\
                                   s_{w,r}^{1}=\{7,1\} 
                              \end{array}\right] \times 2$} &
\multicolumn{1}{c}{$\left[\begin{array}{c}
                                   s_{w,r}^{0}=\{1,7\} \\
                                   s_{w,r}^{1}=\{7,1\} 
                              \end{array}\right] \times 2$} &
\multicolumn{1}{c}{$\left[\begin{array}{c}
                                   s_{w,r}^{0}=\{1,7\} \\
                                   s_{w,r}^{1}=\{7,1\} 
                              \end{array}\right] \times 2$}                              \\
            \bottomrule
    \end{tabular}}
    \vspace{2pt}
    \caption{Model configurations for our focal Transformers. We introduce three configurations Focal-Tiny, Focal-Small and Focal-Base with different model capacities.}
    \label{tab:model_config}
    \vspace{-6mm}
\end{table}

We consider three different network configurations for our focal Transformers. Here, we simply follow the design strategy suggested by previous works~\cite{wang2021pyramid,wu2021cvt,liu2021swin}, though we believe there should be a better configuration specifically for our focal Transformers. Specifically, we use similar design to the Tiny, Small and Base models in Swin Transformer~\cite{liu2021swin}, as shown in Table~\ref{tab:model_config}. Our models take $224 \times 224$ images as inputs and the window partition size is also set to 7 to make our models comparable to the Swin Transformers. For the focal self-attention layer, we introduce two levels, one for fine-grain local attention and one for coarse-grain global attention. Expect for the last stage, the focal region size is consistently set to $13$ for the window partition size of 7, which means that we expand 3 tokens for each window partition. For the last stage, since the whole feature map is $7 \times 7$, the focal region size at level 0 is set to 7, which is sufficient to cover the entire feature map. For the coarse-grain global attention, we set its focal window size same to the window partition size 7, but gradually decrease the focal region size to get $\{7,5,3,1\}$ for the four stages. For the patch embedding layer, the spatial reduction ratio $p_i$ for four stages are all $\{4,2,2,2\}$, while Focal-Base has a higher hidden dimension compared with Focal-Tiny and Focal-Small.
\section{Related work}
% \CL{The first half of this paragraph seems a copy-paste of introduction, we can consider to remove or simplify either one of them.}
\textbf{Vision Transformers}. The Vision Transformer (ViT) was first introduced in~\cite{dosovitskiy2020image}. It applies a standard Transformer encoder, originally developed for NLP~\cite{vaswani2017attention}, to encode image by analogously splitting an image into a sequence of visual tokens. It has demonstrated superior performance to convolutional neural networks (CNNs) such as the ResNet~\cite{he2016deep} on multiple image classification benchmarks, when trained with sufficient data~\cite{dosovitskiy2020image} and careful data augmentation and regularization~\cite{touvron2020training}. These advancements further inspired the applications of transformer to various vision tasks beyond image classification, such as self-supervised learning~\cite{chen2021empirical,caron2021emerging,li2021efficient}, object detection~\cite{carion2020end,zhu2020deformable,zheng2020end,dai2020up} and semantic segmentation~\cite{wang2020max,wang2020end,zheng2021rethinking}. Apart from the downstream tasks, another line of work focus on improving the original vision transformers from different perspectives, such as data-efficient training~\cite{touvron2020training}, improved patch embedding/encoding~\cite{chu2021conditional,yuan2021tokens,han2021transformer}, integrating convolutional projections into transformers~\cite{wu2021cvt,yuan2021incorporating}, multi-scale architectures and efficient self-attention mechanisms for high-resolution vision tasks~\cite{wang2021pyramid,wu2021cvt,liu2021swin,zhang2021multi,chu2021twins}. We refer the readers to~\cite{khan2021transformers,han2020survey,khan2021transformers} for comprehensive surveys. This paper focuses on improving the general performance of vision transformer with the proposed focal self-attention mechanism. In the following, we particularly discussed the most related works regarding attention mechanisms.

\textbf{Efficient global and local self-attention}. 
% Introduce efficient self-attention mechanism for vision tasks. On the other hand, there are many efficient self-attention mechanisms developed in language domain. 
Transformer models usually need to cope with a large number of tokens, such as long documents in NLP and high-resolution images in CV. Recently, various efficient self-attention mechanisms are proposed to overcome the quadratic computational and memory cost in the vanilla self-attention. On one hand, a number of works in both NLP and CV resort to coarse-grained global self-attention by attending the downsampled/summarized tokens, while preserving the long-range interactions~\cite{rae2019compressive,Pappagari2019, wang2021pyramid,wu2021cvt,han2021transformer}. Though this approach can improve the efficiency, it 
loses the detailed context surrounding the query tokens. On the other hand, the local fine-grained attention, \textit{i.e.}, attending neighboring tokens within a constant window size, is another solution for both language~\cite{beltagy2020longformer,zaheer2020big,ainslie2020etc} and vision~\cite{vaswani2021scaling,liu2021swin,zhang2021multi}. In this paper, we argue that both types of attentions are important and the full-attention ViT models indeed have learned both of them, as shown in Fig.~\ref{fig:teaser_fig} left. This is also supported by the recent advanced CNN models~\cite{hu2018squeeze,woo2018cbam,wang2018non,yang2019cross,bello2019attention,cao2019gcnet,srinivas2021bottleneck}, which showed that global attention or interaction can effectively improve the performance. Our proposed focal self-attention is the first to reconcile the global and local self-attention in a single transformer layer.  It can capture both local and global interactions as vanilla full attention but in more efficient and effective way, particularly for high-resolution inputs.
% Finally, please refer to \cite{tay2020efficient,tay2020long,zhang2021multi} for a comprehensive survey and benchmarks of various efficient attention mechanims in NLP and CV applications.

% \textbf{}

\section{Experiments}

\begin{table}[t!]
\begin{minipage}{0.45\linewidth}
\footnotesize
\centering
\setlength{\tabcolsep}{2.1pt}
\resizebox{0.95\linewidth}{!}{
    \begin{tabular}{l|ccc}
    \toprule
    Model & \#Params. & FLOPs & Top-1 (\%) \\
    % \midrule
    % \multicolumn{5}{c}{ImageNet-1K Trained} \\
    \midrule
    ResNet-50~\cite{he2016deep} & 25.0 & 4.1 & 76.2 \\
    DeiT-Small/16~\cite{touvron2020training} & 22.1 & 4.6 & 79.9 \\
    PVT-Small~\cite{wang2021pyramid}  & 24.5 & 3.8 & 79.8 \\
    % CrossViT-Small~\cite{chen2021crossvit} & 224$^2$ & 26.7 & & 81.0 \\
    % LeViT-256~\cite{graham2021levit} & 224$^2$ & 18.9 & 1.1 & 81.6 \\
    ViL-Small~\cite{zhang2021multi}   & 24.6 & 5.1 & 82.0 \\
    CvT-13~\cite{wu2021cvt}        & 20.0 & 4.5 & 81.6 \\
    Swin-Tiny~\cite{liu2021swin}   & 28.3 & 4.5 & 81.2 \\
    \rowcolor{Gray}
    Focal-Tiny~(Ours)               & 29.1 & 4.9 & 82.2 \\
    % Focal-Tiny$^{\dagger}$~(Ours)   & 25.9 & 4.8 & 82.3? \\
    \hline
    ResNet-101~\cite{he2016deep}  & 45.0 & 7.9 & 77.4 \\
    % T2T-ViT-19~\cite{yuan2021tokens} & 224$^2$ &  & & 81.9 \\
    PVT-Medium~\cite{wang2021pyramid} & 44.2 & 6.7 & 81.2 \\
    CvT-21~\cite{wu2021cvt}  & 32.0 & 7.1 & 82.5 \\
    % LeViT-384~\cite{graham2021levit} & 224$^2$ & 39.1 & 2.3 & 82.6 \\
    ViL-Medium~\cite{zhang2021multi}  & 39.7 & 9.1 & 83.3 \\
    Swin-Small~\cite{liu2021swin}    &  49.6 & 8.7 & 83.1 \\  
    \rowcolor{Gray}
    Focal-Small~(Ours) & 51.1 & 9.1 & {83.5} \\
    \hline
    ResNet-152~\cite{he2016deep}  & 60.0 & 11.0 & 78.3 \\
    ViT-Base/16~\cite{dosovitskiy2020image} & 86.6 & 17.6 & 77.9 \\
    DeiT-Base/16~\cite{touvron2020training}& 86.6 & 17.5 & 81.8 \\
    % T2T-ViT-24~\cite{yuan2021tokens} & 224$^2$ & 64.1 & 14.1 & 82.3 \\
    % TNT-Base~\cite{han2021Transformer} & 224$^2$ & 65.6 & 14.1 & 82.8\\
    % CPVT-Base~\cite{chu2021conditional} & 224$^2$ & 88.0 & 17.6 & 82.3 \\
    % CrossViT-Base~\cite{chen2021crossvit} & 224$^2$ & 104.7 & 21.2 & 82.2 \\
    PVT-Large~\cite{wang2021pyramid} & 61.4 & 9.8 & 81.7 \\
    ViL-Base~\cite{zhang2021multi} & 55.7 & 13.4 & 83.2 \\
    Swin-Base~\cite{liu2021swin}   & 87.8 & 15.4 & 83.4 \\  
    % Swin-Base~\cite{liu2021swin}   &  384$^2$ & 87.8 & 15.4 & & 83.4 \\ 
    \rowcolor{Gray}
    Focal-Base~(Ours)  & 89.8 & 16.0 & \textbf{83.8} \\
    % \bottomrule
    % Focal-Base   & 384$^2$ & 89.8 & & & 83.5? \\
    % \midrule
    % \multicolumn{5}{c}{ImageNet-22K Pretrained} \\
    % \midrule
    % % ViT-Large/16~\cite{dosovitskiy2020image} & 384$^2$ \\
    % Swin-Base~\cite{liu2021swin} & 224$^2$  \\    
    % % Swin-Base~\cite{liu2021swin} & 384$^2$  \\    
    % Swin-Large~\cite{liu2021swin} & 224$^2$  \\    
    % % Swin-Large~\cite{liu2021swin} & 384$^2$  \\    
    % Focal-Base  & 224$^2$ &   \\
    % % Focal-Base  & 384$^2$ &   \\
    % Focal-Large  & 224$^2$ &   \\
    % % Focal-Large  & 384$^2$ &   \\
     \bottomrule
    \end{tabular}
    
    }
    \vspace{2pt}
    \caption{Comparison of image classification on ImageNet-1K for different models. Except for ViT-Base/16, all other models are trained and evaluated on $224 \times 224$ resolution.}
    \label{tab:image_classification}
\end{minipage}
\quad
\begin{minipage}{0.52\linewidth}
\footnotesize
\centering
\setlength{\tabcolsep}{2.6pt}
\resizebox{0.98\linewidth}{!}{
    \begin{tabular}{l|l|ll}
    \toprule
     \multirow{2}{*}{Backbone}        & RetinaNet & \multicolumn{2}{c}{Mask R-CNN} \\
     & $AP^{b}$ & $AP^{b}$ & $AP^{m}$\\
    % \midrule
    % \multicolumn{5}{c}{ImageNet-1K Trained} \\
    \midrule
    ResNet-50~\cite{he2016deep} & 36.3 & 38.0 & 34.4 \\
    PVT-Small & 40.4 & 40.4 & 37.8 \\
    ViL-Small~\cite{zhang2021multi} & 41.6 & 41.8 & 38.5 \\
    Swin-Tiny~\cite{liu2021swin} & 42.0 & 43.7 & 39.8 \\
    \rowcolor{Gray}
    Focal-Tiny~(Ours)            & \textbf{43.7}~\scriptsize{\textcolor{blue}{\bf(+1.7)}} & \textbf{44.8}~\scriptsize{\textcolor{blue}{\bf(+1.1)}} & \textbf{41.0}~\scriptsize{\textcolor{blue}{\bf(+1.3)}} \\
    \hline
    ResNet-101~\cite{he2016deep} & 38.5 & 40.4 & 36.4 \\
    ResNeXt101-32x4d~\cite{xie2017aggregated} & 39.9 & 41.9 & 37.5 \\
    PVT-Medium~\cite{wang2021pyramid} & 41.9 & 42.0 & 39.0 \\
    ViL-Medium~\cite{zhang2021multi} & 42.9 & 43.4 & 39.7 \\
    Swin-Small~\cite{liu2021swin}  & 45.0 & 46.5 & 42.1 \\  
    \rowcolor{Gray}
    Focal-Small~(Ours) & \textbf{45.6}~\scriptsize{\textcolor{blue}{\bf(+0.6)}} & \textbf{47.4}~\scriptsize{\textcolor{blue}{\bf(+0.9)}} & \textbf{42.8}~\scriptsize{\textcolor{blue}{\bf(+0.7)}} \\
    \hline
    % ResNet-152~\cite{he2016deep} & 60.0 & 11.0 & 78.3 \\
    ResNeXt101-64x4d~\cite{xie2017aggregated} & 41.0 & 42.8 & 38.4 \\    
    PVT-Large~\cite{wang2021pyramid}  & 42.6 & 42.9 & 39.5 \\
    ViL-Base~\cite{zhang2021multi} & 44.3 & 45.1 & 41.0 \\
    Swin-Base~\cite{liu2021swin} & 45.0 & 46.9 & 42.3 \\  
    \rowcolor{Gray}
    Focal-Base~(Ours) & \textbf{46.3}~\scriptsize{\textcolor{blue}{\bf(+1.3)}} & \textbf{47.8}~\scriptsize{\textcolor{blue}{\bf(+0.9)}} & \textbf{43.2}~\scriptsize{\textcolor{blue}{\bf(+0.9)}} \\
    \bottomrule
    \end{tabular}}
    \vspace{3pt}
    \caption{Comparisons with CNN and Transformer baselines and SoTA methods on COCO object detection. The box mAP ($AP^b$) and mask mAP ($AP^m$) are reported for RetinaNet and Mask R-CNN trained with $1\times$ schedule. More detailed comparisons with $3\times$ schedule are in Table~\ref{tab:object_detection_3x}.}
    \label{tab:object_detection}
\end{minipage}
\end{table}

\subsection{Image classification on ImageNet-1K}

We compare different methods on ImageNet-1K~\cite{deng2009imagenet}. For fair comparison, we follow the training recipes in~\cite{touvron2020training,wang2021pyramid}. All models are trained for 300 epochs with a batch size 1024. The initial learning rate is set to $10^{-3}$ with 20 epochs of linear warm-up starting from $10^{-5}$. For optimization, we use AdamW~\cite{loshchilov2017decoupled} as the optimizer with a cosine learning rate scheduler. The weight decay is set to $0.05$ and the maximal gradient norm is clipped to 5.0. We use the same set of data augmentation and regularization strategies used in~\cite{touvron2020training} after excluding random erasing~\cite{zhong2020random}, repeated augmentation~\cite{berman2019multigrain,hoffer2020augment} and exponential moving average (EMA)~\cite{polyak1992acceleration}. 
The stochastic depth drop rates are set to $0.2$, $0.2$ and $0.3$ for our tiny, small and base models, respectively. During training, we crop images randomly to $224 \times 224$, while a center crop is used during evaluation on the validation set.

In Table~\ref{tab:image_classification}, we summarize the results for baseline models and the current state-of-the-art models on image classification task. We can find our Focal Transformers consistently outperforms other methods with similar model size (\#Params.) and computational complexity (GFLOPs). Specifically, Focal-Tiny improves over the Transformer baseline DeiT-Small/16 by 2.0\%. Meanwhile, using the same model configuration (2-2-6-2) and a few extra parameters and computations, our Focal-Tiny improves over Swin-Tiny by 1.0 point (81.2\% $\rightarrow$ 82.2\%). When we increase the window size from 7 to 14 to match the settings in ViL-Small~\cite{zhang2021multi}, the performance can be further improved to 82.5\%. For small and base models, our Focal Transformers still achieves slightly better performance than the others. Notably, our Focal-Small with 51.1M parameters can reach 83.5\% which is better than all counterpart small and base models using much less parameters. When further increasing the model size, our Focal-Base model achieves 83.8\%, surpassing all other models using comparable parameters and FLOPs. We refer the readers to our appendix for more detailed comparisons.

\begin{table}[!ht]
\begin{center}
\resizebox{\linewidth}{!}{
\setlength{\tabcolsep}{2.1pt}
\footnotesize
\begin{tabular}{l|cc|cccccc|cccccc}
\toprule
\multirow{2}{*}{Backbone} & \#Params & FLOPs &  \multicolumn{6}{c}{RetinaNet 3x schedule + MS} & \multicolumn{6}{c}{Mask R-CNN 3x schedule + MS}\\
 & (M) & (G) & $AP^b$ & $AP^b_{50}$ & $AP^b_{75}$ & $AP_{S}$ & $AP_{M}$ & $AP_{L}$ & $AP^b$ & $AP^b_{50}$ & $AP^b_{75}$ & $AP^m$ & $AP^m_{50}$ & $AP^m_{75}$\\
\midrule
ResNet50~\cite{he2016deep} & 37.7/44.2 & 239/260 & 39.0 & 58.4 & 41.8 & 22.4 & 42.8 & 51.6 & 41.0 & 61.7 & 44.9 & 37.1 & 58.4 & 40.1 \\
PVT-Small\cite{wang2021pyramid} & 34.2/44.1 & 226/245 & 42.2 & 62.7 & 45.0 & 26.2 & 45.2 & 57.2 & 43.0 & 65.3 & 46.9 & 39.9 & 62.5 & 42.8 \\
ViL-Small~\cite{zhang2021multi} & 35.7/45.0 & 252/174 & 42.9 & 63.8 & 45.6 & 27.8 & 46.4 & 56.3 & 43.4 & 64.9 & 47.0 & 39.6 & 62.1 & 42.4 \\
Swin-Tiny~\cite{liu2021swin} & 38.5/47.8 & 245/264 & 45.0 & 65.9 & 48.4 & 29.7 & 48.9 & 58.1 & 46.0 & 68.1 & 50.3 & 41.6 & 65.1 & 44.9 \\
% https://ml.azure.com/job/gcrprojvc1/az-wus2-v100-32gb-3/b060087b7f6e41d500eec6075e724ae4
% https://ml.azure.com/job/msrhyper/msr-lab-dgx2/53c58ee0586717baaeb06e954539ec2d
\rowcolor{Gray}
Focal-Tiny (Ours) & 39.4/48.8 & 265/291 & \textbf{45.5} & \textbf{66.3} & \textbf{48.8} & \textbf{31.2} & \textbf{49.2} & \textbf{58.7} & \textbf{47.2} & \textbf{69.4} & \textbf{51.9} & \textbf{42.7} & \textbf{66.5} & \textbf{45.9} \\
% https://ml.azure.com/job/msrhyper/msr-lab-hyper-dgx2/4ef4f48228c8438937d71a765ba93af8
% https://ml.azure.com/job/msrhyper/msr-lab-dgx2/3decd79753d86cf3169aa284a6090bf9
% Focal-T/14 & - & - &  \\
\midrule
ResNet101~\cite{he2016deep} & 56.7/63.2 & 315/336 & 40.9 & 60.1 & 44.0 & 23.7 & 45.0 & 53.8 & 42.8 & 63.2 & 47.1 & 38.5 & 60.1 & 41.3 \\
ResNeXt101-32x4d~\cite{xie2017aggregated} & 56.4/62.8 & 319/340 & 41.4 & 61.0 & 44.3 & 23.9 & 45.5 & 53.7 & 44.0 & 64.4 & 48.0 & 39.2 & 61.4 & 41.9 \\
PVT-Medium~\cite{wang2021pyramid} & 53.9/63.9 & 283/302 & 43.2 & 63.8 & 46.1 & 27.3 & 46.3 & 58.9 & 44.2 & 66.0 & 48.2 & 40.5 & 63.1 & 43.5 \\
ViL-Medium~\cite{zhang2021multi} & 50.8/60.1 & 339/261 & 43.7 & 64.6 & 46.4 & 27.9 & 47.1 & 56.9 & 44.6 & 66.3 & 48.5 & 40.7 & 63.8 & 43.7 \\
Swin-Small~\cite{liu2021swin} & 59.8/69.1 & 335/354 & 46.4 & 67.0 & 50.1 & 31.0 & 50.1 & 60.3  & 48.5 & 70.2 & 53.5 & 43.3 & 67.3 & 46.6 \\
% https://ml.azure.com/job/msrhyper/msr-lab-dgx2/867dfc757f15c3dd13515fba83b096a5
% https://ml.azure.com/job/msrhyper/msr-lab-dgx2/f3094efd21955f989585a55c63a6f3a1
\rowcolor{Gray}
Focal-Small (Ours) & 61.7/71.2 & 367/401 & \textbf{47.3} & \textbf{67.8} & \textbf{51.0} & \textbf{31.6} & \textbf{50.9} & \textbf{61.1} & \textbf{48.8} & \textbf{70.5} & \textbf{53.6} & \textbf{43.8} & \textbf{67.7} & \textbf{47.2} \\

% 47.2634,67.7992,51.0353,31.5859,50.8644,61.1443
% https://ml.azure.com/runs/swin_Transformer-33e2a88e-well-ray-trainval_retina_with_swin_small-33714287?wsid=/subscriptions/46da6261-2167-4e71-8b0d-f4a45215ce61/resourcegroups/msrhyperVC/workspaces/msrhyper&tid=72f988bf-86f1-41af-91ab-2d7cd011db47#outputsAndLogs
% Focal-S/14 & - & - &  \\
\midrule
ResNeXt101-64x4d~\cite{xie2017aggregated} & 95.5/102 & 473/493 & 41.8 & 61.5 & 44.4 & 25.2 & 45.4 & 54.6 & 44.4 & 64.9 & 48.8 & 39.7 & 61.9 & 42.6 \\
PVT-Large\cite{wang2021pyramid} & 71.1/81.0 & 345/364 & 43.4 & 63.6 & 46.1 & 26.1 & 46.0 & 59.5 & 44.5 & 66.0 & 48.3 & 40.7 & 63.4 & 43.7 \\
ViL-Base~\cite{zhang2021multi} & 66.7/76.1 & 443/365 & 44.7 & 65.5 & 47.6 & 29.9 & 48.0 & 58.1 & 45.7 & 67.2 & 49.9 & 41.3 & 64.4 & 44.5 \\
Swin-Base~\cite{liu2021swin} & 98.4/107 & 477/496 & 45.8 & 66.4 & 49.1 & 29.9 & 49.4 & 60.3 & 48.5 & 69.8 & 53.2 & 43.4 & 66.8 & 46.9 \\
% https://ml.azure.com/job/msrhyper/msr-lab-dgx2/24d03a24deb92bae34180861e79be137
% https://ml.azure.com/job/msrhyper/msr-lab-dgx2/2dbd6318c7572bab2b305c1eb8c0e08e
\rowcolor{Gray}
Focal-Base (Ours) & 100.8/110.0 & 514/533 & \textbf{46.9} & \textbf{67.8} & \textbf{50.3} & \textbf{31.9} & \textbf{50.3} & \textbf{61.5} & \textbf{49.0} & \textbf{70.1} & \textbf{53.6} & \textbf{43.7} & \textbf{67.6} & \textbf{47.0} \\
% 46.9125,67.8018,50.2914,31.8636,50.3402,61.4545
% https://ml.azure.com/job/msrhyper/msr-lab-dgx2/738a9291c2527cd3faad9233b3577d6b
% Focal-B/14 & - & - &  \\
\bottomrule
\end{tabular}}
\end{center}
\vspace{2pt}
\caption{COCO object detection and segmentation results with RetinaNet~\cite{lin2017focal} and Mask R-CNN~\cite{he2016deep}. All models are trained with $3\times$ schedule and multi-scale inputs (MS). The numbers before and after ``/'' at column 2 and 3 are the model size and complexity for RetinaNet and Mask R-CNN, respectively.}
\label{tab:object_detection_3x}
\vspace{-2mm}
\end{table}

\subsection{Object detection and instance segmentation}

We benchmark our models on object detection with COCO~2017~\cite{lin2014microsoft}. The pretrained models are used as visual backbones and then plug into two representative pipelines, RetinaNet~\cite{lin2017focal} and Mask R-CNN~\cite{he2017mask}. All models are trained on the 118k training images and results reported on 5K validation set. We follow the standard to use two training schedules, $1\times$ schedule with 12 epochs and $3\times$ schedule with 36 epochs. For $1\times$ schedule, we resize image's shorter side to $800$ while keeping its longer side no more than 1333. For $3\times$  schedule, we use multi-scale training strategy by randomly resizing its shorter side to the range of $[480,800]$. Considering this higher input resolution, we adaptively increase the focal sizes at four stages to $(15,13,9,7)$, to ensures the focal attention covers more than half of the image region (first two stages) to the whole image ( last two stages). With the focal size increased, the relative position biases are accordingly up-sampled to corresponding sizes using bilinear interpolation. During training, we use AdamW~\cite{loshchilov2017decoupled} for optimization with initial learning rate $10^{-4}$ and weight decay $0.05$. Similarly, we use 0.2, 0.2 and 0.3 stochastic depth drop rates to regularize the training for our tiny, small and base models, respectively. Since Swin Transformer does not report the numbers on RetinaNet, we train it by ourselves using their official code with the same hyper-parameters with our Focal Transformers.

In Table~\ref{tab:object_detection}, we show the performance for both CNN-based models and the current Transformer-based state-of-the-arts methods. The bbox mAP ($AP^b$) and mask mAP ($AP^m$) are reported. Our Focal Transformers outperform the CNN-based models consistently with the gap of 4.8-7.1 points. Compared with the other methods which also use multi-scale Transformer architectures, we still observe substantial gains across all settings and metrics. Particularly, our Focal Transformers brings 0.7-1.7 points of mAP against the current best approach Swin Transformer~\cite{liu2021swin} at comparable settings. Different from the other multi-scale Transformer models, our method can simultaneously enable short-range fine-grain and long-range coarse-grain interactions for each visual token, and thus capture richer visual contexts at each layer for better dense predictions. To have more comprehensive comparisons, we further train them with  $3\times$  schedule and show the detailed numbers for RetinaNet and Mask R-CNN in Table~\ref{tab:object_detection_3x}. For comprehension, we also list the number of parameters and the associated computational cost for each model. As we can see, even for  $3\times$  schedule, our models can still achieve 0.3-1.1 gain over the best Swin Transformer models at comparable settings.

\begin{wraptable}{r}{8.0cm}
\vspace{-3pt}
\begin{minipage}{1.0\linewidth}
\resizebox{1.0\linewidth}{!}{
\centering
\footnotesize
\setlength{\tabcolsep}{2.5pt}
  \begin{tabular}{clcc|lll}
    \toprule
    Method & Backbone & \#Param & FLOPs & $AP^b$ & $AP^b_{50}$ & $AP^b_{75}$ \\
    \midrule	 
    \multirow{3}{*}{\makecell{C. Mask R-CNN~\cite{cai2018cascade}}} 
         & R-50 & 82.0 & 739 & 46.3 & 64.3 & 50.5 \\
         & Swin-T & 85.6 & 742 & 50.5 & 69.3 & 54.9 \\
           \rowcolor{Gray}    \cellcolor{white}   
         & Focal-T & 86.7 & 770 & \textbf{51.5}~\scriptsize{\textcolor{blue}{\bf(+1.0)}} & \textbf{70.6} & \textbf{55.9} \\
        %  https://ml.azure.com/job/msrhyper/msr-lab-hyper-dgx2/4ed0b52ab4058a45e8f246729b2089f7
    \midrule
    \multirow{3}{*}{ATSS~\cite{zhang2020bridging}} 
         & R-50 & 32.1 & 205 & 43.5 & 61.9 & 47.0 \\
         & Swin-T & 35.7 & 212 & 47.2 & 66.5 & 51.3 \\
           \rowcolor{Gray}    \cellcolor{white}   
         & Focal-T & 36.8 & 239 & \textbf{49.5}~\scriptsize{\textcolor{blue}{\bf(+2.3)}} & \textbf{68.8} & \textbf{53.9}\\
        %  https://ml.azure.com/job/msrhyper/msr-lab-dgx2/ad6216db236ab0452ce24830381ecc00
    \midrule
    \multirow{3}{*}{RepPointsV2~\cite{yang2019reppoints}} 
         & R-50 & 43.4 & 431 & 46.5 & 64.6 & 50.3 \\
         & Swin-T & 44.1 & 437 & 50.0 & 68.5 & 54.2 \\
           \rowcolor{Gray}    \cellcolor{white}   
         & Focal-T & 45.4 & 491 & \textbf{51.2}~\scriptsize{\textcolor{blue}{\bf(+1.2)}} & \textbf{70.4} & \textbf{54.9} \\
    \midrule
    \multirow{3}{*}{Sparse R-CNN~\cite{sun2020sparse}} 
         & R-50 & 106.1 & 166 & 44.5 & 63.4 & 48.2 \\
         & Swin-T & 109.7 & 172 & 47.9 & 67.3 & 52.3 \\
    \rowcolor{Gray}    \cellcolor{white}   
         & 
         Focal-T & 110.8 & 196 & \textbf{49.0}~\scriptsize{\textcolor{blue}{\bf(+1.1)}} & \textbf{69.1} & \textbf{53.2} \\
         % https://ml.azure.com/job/msrhyper/msr-lab-dgx2/83ebeb2621447181aeabb6544b470a41
        %  49.0:https://ml.azure.com/job/msrhyper/msr-lab-dgx2/554bd4236d1bd0147ffe2f284a582f51
 \bottomrule
  \end{tabular}}
  \vspace{3pt}
    \caption{Comparison with ResNet-50, Swin-Tiny across different object detection methods. We use Focal-Tiny as the backbone and train all models using $3\times$ schedule.}
    \label{tab:ablation_on_detectors}
\end{minipage}
\vspace{-2mm}
\end{wraptable}

To further verify the effectiveness of our proposed Focal Transformers, we follow~\cite{liu2021swin} to train four different object detectors including Cascade R-CNN~\cite{cai2018cascade}, ATSS~\cite{zhang2020bridging}, RepPoints~\cite{yang2019reppoints} and Sparse R-CNN~\cite{sun2020sparse}. We use Focal-Tiny as the backbone and train all four models using $3\times$ schedule. The box mAPs on COCO validation set are reported in Table~\ref{tab:ablation_on_detectors}. As we can see, our Focal-Tiny exceeds Swin-Tiny by 1.0-2.3 points on all methods. These significant and consistent improvements over different detection methods in addition to RetinaNet and Mask R-CNN suggest that our Focal Transformer can be used as a generic backbone for a variety of object detection methods.

Besides the instance segmentation results above, we further evaluate our model on semantic segmentation, a task that usually requires high-resolution input and long-range interactions. We benchmark our method on ADE20K~\cite{zhou2017scene}. Specifically, we use UperNet~\cite{xiao2018unified} as the segmentation method and our Focal Transformers as the backbone. We train three models with Focal-Tiny, Focal-Small, Focal-Base, respectively. For all models, we use a standard recipe by setting the input size to $512 \times 512$ and train the model for 160k iterations with batch size 16. In Table~\ref{tab:semantic_segmentation}, we show the comparisons to previous works. As we can see, our tiny, small and base models consistently outperforms Swin Transformers with similar size on single-scale and multi-scale mIoUs. 

\vspace{-5pt}
\begin{table}
\begin{minipage}{0.52\linewidth}
\setlength{\tabcolsep}{1.8pt}
\centering
\footnotesize
\resizebox{\linewidth}{!}{
  \begin{tabular}{lcc|cc|cc}
    \toprule
     \multirow{2}{*}{Method}      &   \multirow{2}{*}{\#Param}         & \multirow{2}{*}{FLOPs} & \multicolumn{2}{c|}{mini-val} &  \multicolumn{2}{c}{test-dev} \\
     & &  & $AP^b$ & $AP^m$ & $AP^b$ & $AP^m$\\
    \midrule	 
    X101-64x4d~\cite{xie2017aggregated} & 155M & 1033G & 52.3 & 46.0 & - & -\\
    EfficientNet-D7~\cite{tan2019efficientnet} & 77M & 410G & 54.4 & - & 55.1 & - \\
    GCNet$^*$~\cite{cao2019gcnet} & - & 1041G & 51.8 & 44.7 & 52.3 & 45.4 \\
    ResNeSt-200~\cite{zhang2020resnest} & - & - & 52.5 & - & 53.3 & 47.1 \\
    Copy-paste~\cite{ghiasi2020simple} & 185M & 1440G & 55.9 & 47.2 & 56.0 & 47.4 \\
    BoTNet-200~\cite{srinivas2021bottleneck} & - & - & 49.7 & - & - & - \\
    SpineNet-190~\cite{du2020spinenet} & 164M & 1885G & 52.6 & - & 52.8 & -\\
    CenterNet2~\cite{zhou2021probabilistic} & - & - & - & - & 56.4 & - \\
    \midrule
    % Swin-B~\cite{liu2021swin} & 160M & 1043G & 56.4 & 49.1 \\
    Swin-L~(HTC++)~\cite{liu2021swin} & 284M & 1470G & {57.1} & 49.5 & 57.7 & 50.2 \\
    Swin-L~(DyHead)~\cite{dai2021dynamic} & 213M & 965G & 56.2 & - & - & -\\
    Swin-L$^{\dagger}$~(HTC++)~\cite{liu2021swin} & 284M & - & {58.0} & 50.4 & {58.7} & 51.1 \\
    Swin-L$^{\dagger}$~(DyHead)~\cite{dai2021dynamic} & 213M & - & 58.4 & - & 58.7 & -\\
    Swin-L$^{\dagger}$~(QueryInst)~\cite{fang2021instances} & - & - & 56.1 & - & 56.1 & - \\    
    \midrule
    % Swin-L~(HTC++)~(Our run) & 284M & 1470G & 55.9 & 48.9 & - & -\\
    % Swin-L$^{\dagger}$~(HTC++)~(Our run) & 284M & - & 57.1 & 50.0 & - & - \\    
     \rowcolor{Gray} 
    Focal-L~(HTC++)~(Ours) & 265M & 1165G & {57.0} & {49.9} & - & -\\
     \rowcolor{Gray} 
    Focal-L~(DyHead)~(Ours) & 229M & 1081G & 56.4 & - & - & - \\
     \rowcolor{Gray} 
    Focal-L$^{\dagger}$~(HTC++)~(Ours) & 265M & - & {58.1} & \textbf{50.9} & 58.4 & \textbf{51.3} \\        
     \rowcolor{Gray} 
    Focal-L$^{\dagger}$~(DyHead)~(Ours) & 229M & - & \cocominivalboxmapms & - & \cocotestdevboxmapms & - \\
    % https://ml.azure.com/job/gcrprojvc1/az-wus2-v100-32gb-3/d80e16eaf2acffebf845c3b4a5300492
     \bottomrule
  \end{tabular} 
  }
  \vspace{2pt}
  \caption{Comparison with state-of-the-art methods on COCO object detection and instance segmentation. The numbers are reported on 5K val set and test-dev. Augmented HTC~\cite{chen2019hybrid} (denoted by HTC++) and DyHead~\cite{dai2021dynamic} are used as the detection methods. $^{\dagger}$ means multi-scale evaluation.}
  \label{tab:object_detection_sota}
  \end{minipage}
  \quad
\begin{minipage}{0.47\linewidth}
\setlength{\tabcolsep}{1.8pt}
\centering
\footnotesize
\resizebox{0.99\linewidth}{!}{
  \begin{tabular}{lccc|cc}
    \toprule
    Backbone & Method & \#Param & FLOPs & mIoU & +MS \\
    \midrule	 
    ResNet-101 & DANet~\cite{nam2017dual} & 69M & 1119G & 45.3 & -\\
    % ResNet-101 & DLab.v3+~\cite{} & 63M & 1021G & 44.1 \\
    ResNet-101 & ACNet~\cite{fu2019adaptive} & - & - & 45.9 & -\\
    ResNet-101 & DNL~\cite{yin2020disentangled} & 69M & 1249G & 46.0 & - \\
    % ResNet-101 & OCRNet~\cite{} & 56M & 923G & 45.3 \\
    ResNet-101 & UperNet~\cite{xiao2018unified} & 86M & 1029G & 44.9 & - \\
    \midrule
    HRNet-w48~\cite{SunXLW19} & OCRNet~\cite{yuan2019object} & 71M & 664G & 45.7 & -\\
    ResNeSt-200~\cite{zhang2020resnest} & DLab.v3+~\cite{chen2018encoder} & 88M & 1381G & 48.4 & -\\
    \midrule
    Swin-T~\cite{liu2021swin} & UperNet~\cite{xiao2018unified} & 60M & 945G & 44.5 & 45.8  \\
    Swin-S~\cite{liu2021swin} & UperNet~\cite{xiao2018unified} & 81M & 1038G & 47.6 & 49.5  \\
    Swin-B~\cite{liu2021swin} & UperNet~\cite{xiao2018unified} & 121M & 1188G & 48.1 & 49.7  \\
    Twins-SVT-L~\cite{chu2021twins} & UperNet~\cite{xiao2018unified} & 133M & - & 48.8 & 50.2 \\
    MiT-B5~\cite{xie2021segformer} & SegFormer~\cite{xie2021segformer} & 85M & - & 51.0 & 51.8\\
    ViT-L/16$^{\dagger}$~\cite{dosovitskiy2020image} & SETR~\cite{zheng2020rethinking} & 308M & - & 50.3 & - \\    
    Swin-L$^{\ddagger}$~\cite{liu2021swin} & UperNet~\cite{xiao2018unified} & 234M & 3230G & 52.1 & 53.5 \\
    ViT-L/16$^{\ddagger}$~\cite{dosovitskiy2020image} & Segmenter~\cite{strudel2021segmenter} & 334M & - & 51.8 & 53.6 \\
    Swin-L$^{\ddagger}$~\cite{liu2021swin} & K-Net~\cite{zhang2021knet} & - & - & - & 54.3 \\
    Swin-L$^{\ddagger}$~\cite{liu2021swin} & PatchDiverse~\cite{gong2021vision} & 234M & - & 53.1 & 54.4 \\
    VOLO-D5~\cite{yuan2021volo} & UperNet~\cite{xiao2018unified} & - & - & - & 54.3 \\
    \midrule
     \rowcolor{Gray} 
    Focal-T~(Ours) & UperNet~\cite{xiao2018unified} & 62M & 998G & 45.8 & 47.0 \\
     \rowcolor{Gray} 
    Focal-S~(Ours) & UperNet~\cite{xiao2018unified} & 85M & 1130G & 48.0 & 50.0\\     
     \rowcolor{Gray} 
    Focal-B~(Ours) & UperNet~\cite{xiao2018unified} & 126M & 1354G & 49.0 & 50.5 \\      
     \rowcolor{Gray} 
    Focal-L$^{\ddagger}$~(Ours) & UperNet~\cite{xiao2018unified} & 240M & 3376G & \adevalss & \adevalms  \\   
    \bottomrule
    % https://ml.azure.com/job/msrhyper/msr-lab-dgx2/d853287f8ad3b30a490102d4c045675e
    % finetune with 2560x640: https://ml.azure.com/job/msrhyper/msr-lab-hyper-dgx2/659b3299967b1994be5ff7ca3cdcbe6f
    % eval with 2560x640: https://ml.azure.com/job/msrhyper/az-wus3-a100-2/ba71604a81dce8f170c2d114e49376e4
  \end{tabular} 
  }
  \vspace{2pt}
  \caption{Comparison with SoTA methods for semantic segmentation on ADE20K~\cite{zhou2017scene} val set. Both single- and multi-scale evaluations are reported at the last two columns. $^{\ddagger}$ means pretrained on ImageNet-22K.}
  \label{tab:semantic_segmentation}
  \end{minipage}  
\end{table}

\subsection{Comparing with system-level SoTA methods}
To compare with SoTA at the system level, we build Focal Large model by increasing the hidden dimension in Focal-Base from 128 to 196 while keeping all the others the same, similar to Swin Transformers. 
To achieve the best performance, the common practice is to pretrain on ImageNet-22K and then transfer the model to down stream tasks~\cite{wu2021cvt,liu2021swin}. However, due to the limited resources, we partially initialize our model with the pretrained Swin Transformer checkpoint\footnote{Pretrained models are available at https://github.com/microsoft/Swin-Transformer}, considering our network architecture is similar with Swin Transformers except for the window shift and focal self-attention. 
Specifically, we reuse the parameters in Swin-Large model but remove the window shift operation and randomly initialize our own window pooling layer in Eq.~\eqref{eq:pool} and local-to-global relative position bias in Eq.~\eqref{eq:fsa}. Then, we finetune our model on ImageNet-1K to learn the focal-specific parameters. The resulting model is used as the backbone and further finetuned on object detection and semantic segmentation tasks.

\paragraph{Comparison with SoTA detection systems.}
For object detection on COCO, we first follow Swin Transformer to also use HTC~\cite{chen2019hybrid} as the detection method in that it reported SoTA performance on COCO detection when using Swin Transformer as the backbone. For fair comparison, we also use soft-NMS~\cite{bodla2017soft}, instaboost~\cite{fang2019instaboost} and a multi-scale training strategy with shorter side in range $[400, 1400]$ while the longer side is no more than $1600$. We train the model using AdamW~\cite{loshchilov2017decoupled} with base learning rate 1e-4 and weight decay 0.1. The model is trained using standard $3\times$ schedule. The box and mask mAPs on COCO validation set and test-dev are reported in Table~\ref{tab:object_detection_sota}. We show both single-scale evaluation and multi-scale evaluation results.
% For single-scale, we resize the input images to $(1400, 2100)$, and use $(1200, 1800), (1300, 1950), (1400, 2100), (1450, 2200), (1500, 2250), (1600, 2400)$ for our multi-scale evaluation. 
% As a reference, we also train HTC object detector with Swin-Large using the same regime as our Focal-Large model. 
Our Focal-Large model with multi-scale test achieve 58.1 box mAP and 50.9 mask mAP on mini-val set, which are better than the reported numbers for Swin-Large in~\cite{liu2021swin}. When evaluating our model on the test-dev set, it achieves 58.4 box mAP and 51.3 mask mAP, which is slightly better than Swin Transformer. Note that because our model does not include global self-attention layer used in Swin Transformer at the last stage, it has smaller model size and fewer FLOPs. 
More recently, DyHead~\cite{dai2021dynamic} achieves new SoTA on COCO, when combined with Swin-Large. We replace the Swin-Large model with our Focal-Large model, and use the same $2\times$  training schedule as in~\cite{dai2021dynamic}. We report the box mAPs for both mini-val and test-dev. Our Focal-Large clearly bring substantial improvements over both metrics, reaching new SoTA on both metrics.

% Since the training script for Swin-Large with HTC is not publicly available, there is a small gap between our reproduced numbers and those reported in~\cite{liu2021swin}. Comparing with the numbers from our reproduced Swin-Large model, our Focal-Large model clearly achieved better performance on both box and mask mAP. 

\paragraph{Comparison with SoTA semantic segmentation systems.}
We further use the pretrained Focal-Large model as the backbone for semantic segmentation. We follow the same setting as in~\cite{liu2021swin}. Specifically, we use input image size $640\times640$ and train the model for 160k iterations with a batch size of 16. We set the initial learning to 6e-5 and use a polynomial learning rate decay. The weight decay is set to $0.01$. For multi-scale evaluation, we use the same scaling ratios $[0.5,0.75,1.0,1.25,1.5,1.75]$ as in previous works. In Table~\ref{tab:semantic_segmentation}, we see that our Focal-Large achieves significantly better performance than Swin-Large. In both single-scale and multi-scale evaluation, Focal-Large has more than 1 point mIoU improvement, which presents a new SoTA for semantic segmentation on ADE20K. These encouraging results verify the effectiveness of our proposed focal self-attention mechanism in capturing long-range dependencies required by dense visual prediction tasks.

\begin{table}
\begin{minipage}{0.48\linewidth}
\setlength{\tabcolsep}{1.8pt}
\centering
\footnotesize
\resizebox{0.98\linewidth}{!}{
  \begin{tabular}{cc|cccc}
    \toprule
    Model        & W-Size &  FLOPs & Top-1~(\%) & $AP^b$ & $AP^m$ \\
    \midrule
    \multirow{2}{*}{Swin-Tiny}   & 7  & 4.5 & 81.2 & 43.7 & 39.8  \\
                                 & 14 & 4.9 & 82.1 & 44.0 & 40.5  \\
    \midrule
    \multirow{2}{*}{Focal-Tiny}  & 7  & 4.9 & 82.2 & 44.9 & 41.1  \\
                                 & 14 & 5.2 & 82.3 & 45.5 & 41.5  \\
                                  \bottomrule
  \end{tabular}}
  \vspace{2pt}
\captionof{table}{Impact of different window sizes (W-Size). We alter the default size 7 to 14 and observe consistent improvements for both methods.}
\vspace{-3pt}
\label{tab:ablation_on_window_size}  
\end{minipage}\hfill
\begin{minipage}{0.48\linewidth}
\setlength{\tabcolsep}{2.0pt}
\centering
\footnotesize
\resizebox{0.86\linewidth}{!}{
  \begin{tabular}{cc|ccc}
    \toprule
    Model & W-Shift & Top-1 (\%) & $AP^{b}$ & $AP^{m}$\\
    \midrule	 
    \multirow{2}{*}{Swin-Tiny}   & -          & 80.2 & 38.8 & 36.4 \\
                                 & \checkmark & 81.2 & 43.7 & 39.8 \\
    \midrule
    \multirow{2}{*}{Focal-Tiny}  & -          & 82.2 & 44.8 & 41.0 \\
                                 & \checkmark & 81.9 & 44.9 & 41.1\\
                                 \bottomrule
  \end{tabular}}
  \vspace{2pt}
  \captionof{table}{Impact of window shift (W-Shift) on Swin Transformer and Focal Transformer. Tiny models are used.}
  \vspace{-3pt}
  \label{tab:window_shift}  
\end{minipage}
\end{table}

\begin{table}[t]
\begin{minipage}{0.48\linewidth}
\centering
\footnotesize
  \includegraphics[width=1.0\textwidth]{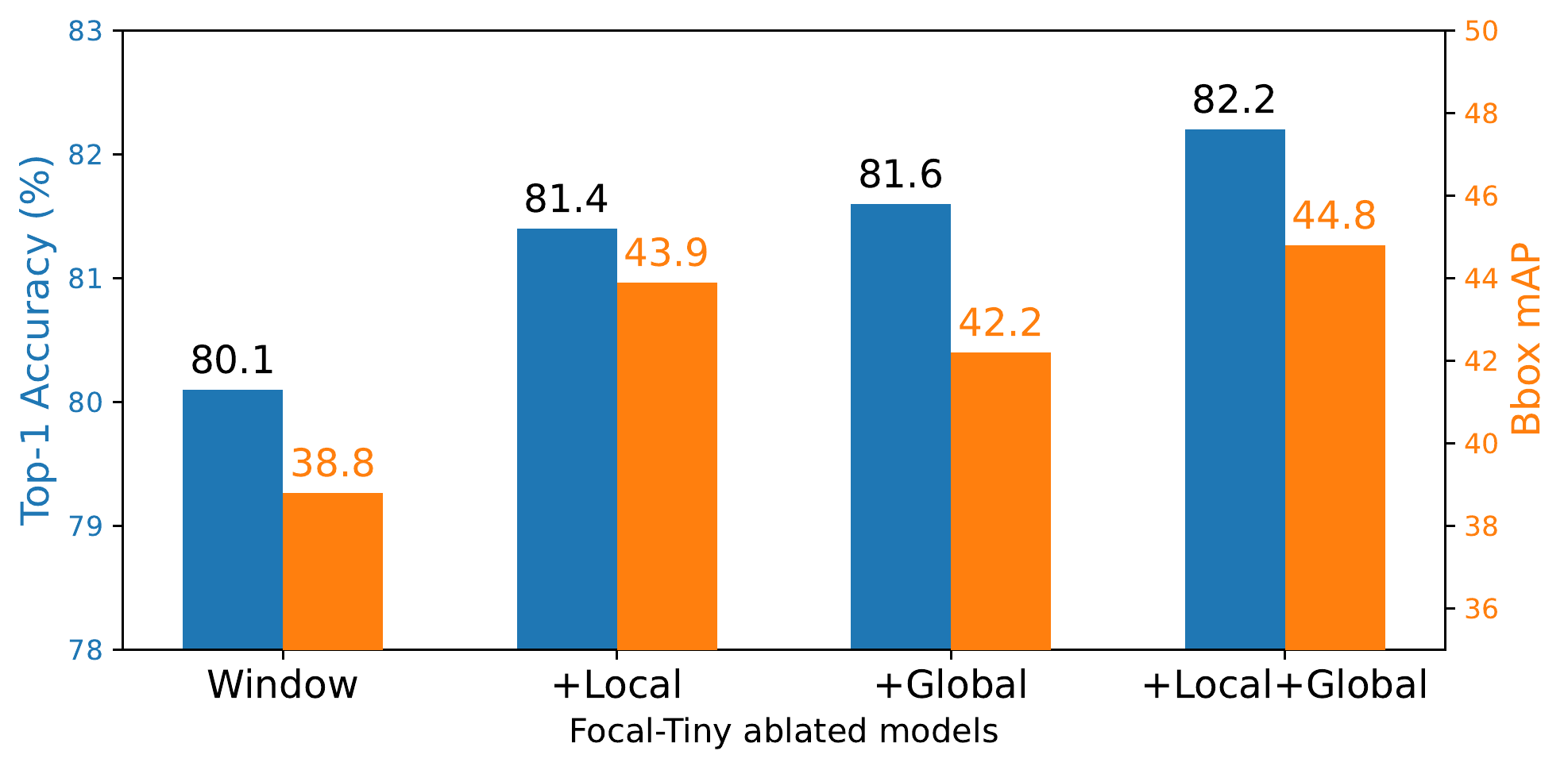}
  \captionof{figure}{Ablating Focal-Tiny model by adding local, global and both interactions, respectively. Blue bars are for image classification and orange bars indicate object detection performance. Both local and global interactions are essential to obtain good performance. Better viewed in color.}
  \label{fig:local_global_ablation}
\end{minipage}
\hfill
\begin{minipage}{0.5\linewidth}
\setlength{\tabcolsep}{1.8pt}
\centering
\footnotesize
\resizebox{0.99\linewidth}{!}{
  \begin{tabular}{cc|ccccc}
    \toprule
    Depths & Model & \#Params. &  FLOPs & Top-1~(\%) & $AP^b$ & $AP^m$ \\
    \midrule
    \multirow{2}{*}{2-2-2-2}   & Swin  & 21.2 & 3.1 & 78.7 & 38.2 & 35.7  \\
      & Focal  & 21.7 & 3.4 & 79.9 & 40.5 & 37.6  \\
      \midrule
    
    \multirow{2}{*}{2-2-4-2}                             & Swin & 24.7 & 3.8 & 80.2 & 41.2 & 38.1  \\
                                 & Focal & 25.4 & 4.1 & \cellcolor{gray!25}{81.4} & 43.3 & \cellcolor{gray!25}{39.8}  \\    
      \midrule
    \multirow{2}{*}{2-2-6-2}                             & Swin & 28.3 & 4.5 & \cellcolor{gray!25}{81.2} & 43.7 & \cellcolor{gray!25}{39.8}  \\                      
                                 & Focal & 29.1 & 4.9 & 82.2 & 44.8 & 41.0  \\ 
                              \bottomrule
  \end{tabular}}
  \vspace{4pt}
  \captionof{table}{Impact of the change of model depth. We gradually reduce the number of transformer layers at the third stage from original 6 to 4 and further 2. It apparently hurts the performance but our Focal Transformers has much slower drop rate than Swin Transformer.}
  \label{tab:performance_with_depth}
\end{minipage}
% \begin{minipage}{0.33\linewidth}
% \setlength{\tabcolsep}{1.8pt}
% \centering
% \footnotesize
% \resizebox{0.98\linewidth}{!}{
%   \begin{tabular}{cc|cccc}
%     \toprule
%     Model        & W-Size &  FLOPs & Top-1~(\%) & $AP^b$ & $AP^m$ \\
%     \midrule
%     \multirow{2}{*}{Swin-Tiny}   & 7  & 4.5 & 81.2 & 43.7 & 39.8  \\
%                                  & 14 & 4.9 & 82.1 & 44.0 & 40.5  \\
%     \midrule
%     \multirow{2}{*}{Focal-Tiny}  & 7  & 4.9 & 81.9? & 44.9 & 41.1  \\
%                                  & 14 & 5.2 & 82.3? & 45.5 & 41.5  \\
%   \end{tabular}}
% \end{minipage}
\quad

\end{table}
\vspace{-3pt}

\subsection{Ablation studies}

% Above we have shown the superior performance of our Focal Transformer. Here 
We conduct ablation studies to inspect the model's capacity from different aspects. Focal-Tiny is considered on both image classification and object detection tasks.
% and have the following main observations.

\textbf{Effect of varying window size}. Above we have demonstrated that both short- and long-range interactions are necessary. Based on this, a natural question is that whether increasing the window size can further help the model learning giving an enlarged receptive field. In  Table~\ref{tab:ablation_on_window_size}, we show the performance of Swin-Tiny and Focal-Tiny with window size 7 and 14. Clearly, a larger window size brings gain for both methods on all three metrics and our Focal-Tiny model consistently outperforms Swin-Tiny using both window sizes. Comparing the second and third row, we find our model beats Swin even using much smaller window size (7 \textit{v.s.} 14). We suspect the long-range interactions in our model is the source of this gain.

\textbf{The necessity of window shift}. In~\cite{liu2021swin}, the authors proposed window shift operations to enable the cross-window interactions between two successive layers. In contrast, the visual tokens in our Focal Transformer can always communicate with those in other windows at both fine- and coarse-grain. A natural question is whether adding the window shift to our Focal Transformers can further lead to improvements. To investigate this, we remove the window shift from Swin Transformer while add it to our Focal Transformer. As shown in Table~\ref{tab:window_shift}, Swin Transformer shows a severe degradation after removing the window shift. However, our Focal Transformer is even hurt on classification task. These results indicate that the window shift is not a necessary ingredient in our model. As such, our model can get rid of the constraint in Swin Transformer that there should be an even number of layers in each stage for the alternative window shift operation.

\textbf{Contributions of short- and long-range interaction}. We attempt to factorize the effect of short-range fine-grain and long-range coarse-grain interactions in our Focal Transformers. We ablate the original Focal-Tiny model to: a) Focal-Tiny-Window merely performing attention inside each window; b) Focal-Tiny-Local attending the additional fine-grain surrounding tokens and c) Focal-Tiny-Global attending the extra coarse-grain squeezed tokens. We train them using the same setting as Focal-Tiny and report their performance on image classification and object detection using Mask R-CNN $1\times$ schedule. As we can see from Fig.~\ref{fig:local_global_ablation}, Focal-Tiny-Window suffers from significant drop on both image classification (82.2$\rightarrow$80.1) and object detection (44.8$\rightarrow$38.3). This is expected since the communication across windows are totally cut off at each Transformer layer. After we enable either the local fine-grain or global coarse-grain interactions (middle two columns), we observe significant jumps. Though they prompt richer interactions from different paths, finally both of them enable the model to capture more contextual information. When we combine them together, we observe further improvements on both tasks. This implies that these two type of interactions are complementary to each other and both of them should be enabled in our model. Another observation is that adding long-range tokens can bring more relative improvement for image classification than object detection and vice versa for local tokens. We suspect that dense predictions like object detection more rely on fine-grained local context while image classification favors more the global information.

\textbf{Model capacity against model depth}. Considering our focal attention prompts local and global interactions at each Transformer layer, one question is that whether it needs less number of layers to obtain similar modeling capacity as those without global interactions. To answer this, we conduct the experiments by reducing the number of Transformer layers at stage 3 in Swin-Tiny and Focal-Tiny from the original 6 to 4 and 2. In Table~\ref{tab:performance_with_depth}, we show the performance and model complexity for each variant. First, we can find our model outperforms Swin model consistently with the same depth. More importantly, using two less layers, our model achieves comparable performance to Swin Transformer. Particularly, Focal-Tiny with 4 layers achieves 81.4 on image classification which is even better than original Swin-Tiny model with 6 layers (highlighted in gray cells). Though we do not explore different architectures for our Focal Transformer, these results suggest that we can potential find even more efficient \emph{and} effective architectures.

\section{Conclusion}
In this paper, we have presented focal self-attention to enable efficient local-global interactions in vision Transformers. Different from previous works, it performs the local self-attention at fine-grain and global self-attention at coarse-grain, which results in an effective way to capture richer context in both short and long-range at a reasonable cost. By plugging it into a multi-scale transformer architecture, we propose Focal Transformers, which demonstrates its superiority over the SoTA methods on both image classification, object detection and segmentation. With these extensive experimental results, the proposed focal attention is shown as a generic approach for modeling local-global interactions in vision Transformers for various vision tasks.

\textbf{Limitations and future work}. Though extensive experimental results showed that our focal self-attention can significantly boost the performance on both image classification and dense prediction tasks, it does introduce extra computational and memory cost, since each query token needs to attend the coarsened global tokens in addition to the local tokens. Developing some practical or methodological techniques to reduce the cost would be necessary to make it more applicable in realistic scenarios. Our ablation study on the number of used Transformer layers in Table~\ref{tab:performance_with_depth} indeed shed light on the potential way to reduce the cost via reducing the number of transformer layers. However, we merely scratched the surface and further study on this aspect is still needed. In Focal Transformers, we chose multi-scale architecture as the base so that it can work for high-resolution prediction tasks. However, we believe our focal attention mechanism is also applicable to monolithic vision Transformers and Transformers in both vision and language domain. We leave this as a promising direction to further explore in the future.
{\small
\bibliographystyle{plain}
\bibliography{neurips.bib}
}

% \section*{References}

% References follow the acknowledgments. Use unnumbered first-level heading for
% the references. Any choice of citation style is acceptable as long as you are
% consistent. It is permissible to reduce the font size to \verb+small+ (9 point)
% when listing the references.
% Note that the Reference section does not count towards the page limit.
% \medskip

% {
% \small

% [1] Alexander, J.A.\ \& Mozer, M.C.\ (1995) Template-based algorithms for
% connectionist rule extraction. In G.\ Tesauro, D.S.\ Touretzky and T.K.\ Leen
% (eds.), {\it Advances in Neural Information Processing Systems 7},
% pp.\ 609--616. Cambridge, MA: MIT Press.

% [2] Bower, J.M.\ \& Beeman, D.\ (1995) {\it The Book of GENESIS: Exploring
%   Realistic Neural Models with the GEneral NEural SImulation System.}  New York:
% TELOS/Springer--Verlag.

% [3] Hasselmo, M.E., Schnell, E.\ \& Barkai, E.\ (1995) Dynamics of learning and
% recall at excitatory recurrent synapses and cholinergic modulation in rat
% hippocampal region CA3. {\it Journal of Neuroscience} {\bf 15}(7):5249-5262.
% }

%%%%%%%%%%%%%%%%%%%%%%%%%%%%%%%%%%%%%%%%%%%%%%%%%%%%%%%%%%%%

%%%%%%%%%%%%%%%%%%%%%%%%%%%%%%%%%%%%%%%%%%%%%%%%%%%%%%%%%%%%
\newpage
\appendix

\section{Appendix}

\subsection{Image classification}

We present the exhaustive comparison with previous works in Table~\ref{tab:full_image_classification}. We compare our method with both CNN-based and Transformer-based methods. We categorize different methods into groups based on two properties:
\begin{itemize}[noitemsep,topsep=0pt,leftmargin=*]
    \item \textbf{Scale} -- the scale of feature maps in a model. It can be either a single-scale or multi-scale. In single-scale models, all feature maps have the same size across different stages. For multi-scale models, there are usually feature maps with different resolutions with the proceeding stages.
    \item \textbf{Locality} -- the locality of operations in a model. It can be either global or local. Local operations can be a convoluional layer in CNN models or a transformer layer which conducts local self-attention. However, global operations such as the standard self-attention, produce the output feature map by gather information from all inputs.
\end{itemize}

Based on this criterion, all CNN models are natural multi-scale because their feature map sizes gradually decrease at different stages. Recently, a number of works attempt to integrate the global operations into CNNs by introducing squeeze-and-excitation (SE) layer~\cite{hu2018squeeze}, channel-wise attention layer~\cite{woo2018cbam} and even self-attention layer~\cite{bello2019attention,srinivas2021bottleneck}. As we can see, the combination of local and global operations significantly improve the performance for image classification. Particularly, BotNet-S1-110 achieves 82.8 top-1 accuracy with moderate number of parameters (61.6M). 

On the contrary, Transformers~\cite{vaswani2017attention} by nature performs global self-attention by which each visual token can interact with all others. Even without multi-scale design as in CNNs, a number of Transformer-based works such as TNT~\cite{han2021transformer}, DeepViT~\cite{zhou2021deepvit} and CaiT~\cite{touvron2021going} achieve superior performance to CNN models with comparable model size and computational cost. To accommodate the high resolution feature maps, some recent works replace global self-attention with more efficient local self-attention and demonstrate comparable performance on image classification while much promising results on dense prediction tasks such as object detection and semantic segmentation~\cite{liu2021swin}.

In this paper, we present focal attention which is the first to combine global self-attention and local self-attention in an efficient way. Replacing either the global self-attention or the local self-attention with our focal self-attention, we achieve better performance than both. These results along with the CNN models augmented by local and global computations demonstrate that combining local and global interactions are more effective than either of them. In the table, we also report the speed for different methods. Using the same script provided by~\cite{liu2021swin}, we run the test on a single Tesla-V100 with batch size 64. Accordingly, our Focal Transformer has slower running speed though it has similar FLOPs as Swin Transformer. This is mainly due to two reasons: 1) we introduce the global coarse-grain attention, and it introduces the extra computations; 2) though we conduct our focal attention on the windows, we swill observe that extracting the surrounding tokens around local windows and the global tokens across the feature map are time-consuming.

\begin{table}[!ht]
\footnotesize
\centering
\resizebox{0.88\linewidth}{!}{
    \begin{tabular}{cccl|ccc}
    \toprule
    Architecture & Scale & Locality & Model & \#Params.~(M) & FLOPs~(G) & Top-1 (\%) \\
    % \midrule
    % \multicolumn{5}{c}{ImageNet-1K Trained} \\
    \midrule
    \multirow{13}{*}{{\makecell{Convolutional \\ Neural Network}}}  & \multirow{13}{*}{Multiple} & \multirow{3}{*}{Local} & ResNet-50~\cite{he2016deep} & 25.0 & 4.1 & 76.2 \\
        & & & ResNet-101~\cite{he2016deep} & 45.0 & 7.9 &77.4 \\
        & & & ResNet-152~\cite{he2016deep} & 60.0 & 11.0 &78.3\\
        \cmidrule{3-7}
        & & \multirow{10}{*}{\makecell{Local \\ +Global}} & SE-ResNet-50~\cite{hu2018squeeze} & 28.1 & 8.2 &77.5 \\
        & & & SE-ResNet-101~\cite{hu2018squeeze} & 49.3 & 15.6 &78.4 \\
        & & & SE-ResNet-152~\cite{hu2018squeeze} & 66.8 & 23.1 &78.9 \\
        \cmidrule{4-7}
        & & & CBAM-ResNet-50~\cite{woo2018cbam} & 28.1 & 3.9 &77.3 \\
        & & & CBAM-ResNet-101~\cite{woo2018cbam} & 49.3 & 7.6 &78.5 \\
        \cmidrule{4-7}
        % & & & Non-Local \\
        & & & AttAug-ResNet-50~\cite{bello2019attention} & 25.8 & 8.3 &77.7 \\
        & & & AttAug-ResNet-101~\cite{bello2019attention} & 45.4 & 16.1  &78.1 \\
        & & & AttAug-ResNet-152~\cite{bello2019attention} & 61.6 & 23.8  &79.1 \\
        % \cmidrule{4-7}
        % & & & GCNet-ResNet-50~\cite{cao2019gcnet} & 28.1 & 3.87 & 77.70 \\
        \cmidrule{4-7}
        & & & BotNet-S1-59~\cite{srinivas2021bottleneck} & 33.5 & 7.3  &81.7 \\
        & & & BotNet-S1-110~\cite{srinivas2021bottleneck} & 54.7 & 10.9 &82.8 \\    
    \midrule
    % Transformer $\rightarrow$ CNN     
    % & & & LeViT-256~\cite{graham2021levit} & 18.9 & 1.1 & 81.6 \\
    % & & & LeViT-384~\cite{graham2021levit} & 39.1 & 2.3 & 82.6 \\
    % \midrule
    \multirow{32}{*}{{\makecell{Transformer}}} & \multirow{14}{*}{{Single}} & \multirow{14}{*}{{Global}} &  ViT-B/16~\cite{dosovitskiy2020image} & 86.6 & 17.6 & 77.9 \\
        & & & ViT-L/16~\cite{dosovitskiy2020image} & 307 & 190.7  &76.5 \\
        \cmidrule{4-7}
      & & & DeiT-S/16~\cite{touvron2020training} & 22.0 & 4.6 & 79.9 \\
      & & & DeiT-B/16~\cite{touvron2020training} & 86.6 & 17.5 & 81.8 \\
      \cmidrule{4-7}
    & & & TNT-S~\cite{han2021transformer} & 23.8 & 5.2 & 81.3 \\
    & & & TNT-B~\cite{han2021transformer} & 65.6 & 14.1 & 82.8\\
    \cmidrule{4-7}
    % & & & CPVT-T~\cite{chu2021conditional} & 6.0 & - & 74.9 \\
    & & & CPVT-S~\cite{chu2021conditional} & 23.0 & 4.6 & 81.5 \\    
    & & & CPVT-B~\cite{chu2021conditional} &  88.0 & 17.6 & 82.3 \\   
    \cmidrule{4-7}
    & & & DeepViT-S~\cite{zhou2021deepvit} & 27.0 & 6.2 & 82.3 \\  
    & & & DeepViT-L~\cite{zhou2021deepvit} & 55.0 & 12.5 & 83.1 \\  
    \cmidrule{4-7}
    & & & CaiT-S36~\cite{touvron2021going} & 68.0 & 13.9 & 83.3 \\    
    \cmidrule{4-7}
    & & & LeViT-256~\cite{graham2021levit} & 18.9 & 1.1 & 81.6 \\
    & & & LeViT-384~\cite{graham2021levit} & 39.1 & 2.3 & 82.6 \\    
    \cmidrule{2-7}
    & \multirow{21}{*}{{Multiple}} & \multirow{9}{*}{{Global}} & T2T-ViT-19~\cite{yuan2021tokens} & 39.2 & 8.9 &81.9 \\
    & & & T2T-ViT-24~\cite{yuan2021tokens} &  64.1 & 14.1 & 82.3 \\
    \cmidrule{4-7}
    & & & CrossViT-S~\cite{chen2021crossvit} & 26.7 & 5.6 & 81.0 \\
    & & & CrossViT-B~\cite{chen2021crossvit} & 104.7 & 21.2 &82.2 \\   
    \cmidrule{4-7}
    & & & PVT-S~\cite{wang2021pyramid} & 24.5 & 3.8 &79.8 \\
    & & & PVT-M~\cite{wang2021pyramid} & 44.2 & 6.7 &81.2 \\
    & & & PVT-L~\cite{wang2021pyramid} & 61.4 & 9.8 &81.7 \\
    \cmidrule{4-7}
    & & & CvT-13~\cite{wu2021cvt} & 20.0 & 4.5 & 81.6 \\
    & & & CvT-21~\cite{wu2021cvt} & 32.0 & 7.1 & 82.5 \\    
    \cmidrule{3-7}
    &  &  & ViL-S~\cite{zhang2021multi}  &  24.6 & 5.1 & 82.0 \\
    & & \multirow{5}{*}{{Local}} & ViL-M~\cite{zhang2021multi} & 39.7 & 9.1 & 83.3 \\
    & & & ViL-B~\cite{zhang2021multi} & 55.7 & 13.4 & 83.2 \\
    \cmidrule{4-7}
    & & & Swin-T~\cite{liu2021swin}     & 28.3 & 4.5  & 81.2 \\
    & & & Swin-S~\cite{liu2021swin}   &  49.6 & 8.7  & 83.1 \\ 
    & & & Swin-B~\cite{liu2021swin}   &  87.8 & 15.4  &  83.4 \\
    \cmidrule{3-7}
    & & \multirow{6}{*}{{\makecell{Local \\ +Global}}} 
        & Twins-SVT-S~\cite{chu2021twins} & 24.0 & 2.8 & 81.3 \\
    & & & Twins-SVT-B~\cite{chu2021twins} & 56.0 & 8.3 & 83.1 \\
    & & & Twins-SVT-L~\cite{chu2021twins} & 99.2 & 14.8 & 83.3 \\
    \cmidrule{4-7}
    & & & Focal-T~(Ours)        & 29.1 & 4.9 & 82.2 \\
    & & & Focal-S~(Ours)        & 51.1 & 9.1 & 83.5 \\
    & & & Focal-B~(Ours)        & 89.8 & 16.0 & \textbf{83.8} \\
    % Swin-Base~\cite{liu2021swin}   &  384$^2$ & 87.8 & 15.4 & & 83.4 \\      
    % Focal-Base   & 384$^2$ & 89.8 & & & 83.5? \\
    % \midrule
    % \multicolumn{5}{c}{ImageNet-22K Pretrained} \\
    % \midrule
    % % ViT-Large/16~\cite{dosovitskiy2020image} & 384$^2$ \\
    % Swin-Base~\cite{liu2021swin} & 224$^2$  \\    
    % % Swin-Base~\cite{liu2021swin} & 384$^2$  \\    
    % Swin-Large~\cite{liu2021swin} & 224$^2$  \\    
    % % Swin-Large~\cite{liu2021swin} & 384$^2$  \\    
    % Focal-Base  & 224$^2$ &   \\
    % % Focal-Base  & 384$^2$ &   \\
    % Focal-Large  & 224$^2$ &   \\
    % % Focal-Large  & 384$^2$ &   \\
    \bottomrule
    \end{tabular}
    }
    \caption{Full comparison of image classification on ImageNet-1k for different model architectures. We split the methods into two super-groups which use CNNs or Transformers as the main skeleton. Note that they are inclusive to each other in some methods.}
    \vspace{-10pt}
    \label{tab:full_image_classification}
\end{table}

\subsection{Object detection and segmentation}

For completeness, we report the full metrics for RetinaNet and Mask R-CNN trained with 1x schedule in Table~\ref{tab:object_detection_1x}. As we can see, our Focal Transformers consistently outperform previous works including the state-of-the-art Swin Transformers on all metrics. We observe that our models trained with 1x schedule generally have more gain against the previous best models than 3x schedule (+1.2 \textit{v.s.} +0.8 and +1.0 \textit{v.s.} +0.7 box mAP for RetinaNet and Mask R-CNN, respectively). This indicates that our models have faster learning convergences compared with previous works. Compared with the local-attention based methods, \textit{e.g.}, Swin Transformer, integrating the long-range interactions can help capture more visual dependencies and thus help the model to learn faster.

\begin{table}[t]
\begin{center}
\resizebox{0.98\linewidth}{!}{
\setlength{\tabcolsep}{2.1pt}
\footnotesize
\begin{tabular}{l|cc|cccccc|cccccc}
\toprule
\multirow{2}{*}{Backbone} & \#Params & FLOPs &  \multicolumn{6}{c}{RetinaNet 1x schedule} & \multicolumn{6}{c}{Mask R-CNN 1x schedule}\\
 & (M) & (G) & $AP^b$ & $AP^b_{50}$ & $AP^b_{75}$ & $AP_{S}$ & $AP_{M}$ & $AP_{L}$ & $AP^b$ & $AP^b_{50}$ & $AP^b_{75}$ & $AP^m$ & $AP^m_{50}$ & $AP^m_{75}$\\
\midrule
ResNet50~\cite{he2016deep} & 37.7/44.2 & 239/260 & 36.3 & 55.3 & 38.6 & 19.3 & 40.0 & 48.8 & 38.0 & 58.6 & 41.4 & 34.4 & 55.1 & 36.7 \\
PVT-Small\cite{wang2021pyramid} & 34.2/44.1 & 226/245 & 40.4 & 61.3 & 43.0 & 25.0 & 42.9 & 55.7 & 40.4 & 62.9 & 43.8 & 37.8 & 60.1 & 40.3 \\
ViL-Small~\cite{zhang2021multi} & 35.7/45.0 & 252/174 & 41.6 & 62.5 & 44.1 & 24.9 & 44.6 & 56.2 & 41.8 & 64.1 & 45.1 & 38.5 & 61.1 & 41.4 \\
Swin-Tiny~\cite{liu2021swin} & 38.5/47.8 & 245/264 & 42.0 & 63.0 & 44.7 & 26.6 & 45.8 & 55.7 & 43.7 & 66.6 & 47.7 & 39.8 & 63.3 & 42.7 \\
% https://ml.azure.com/job/gcrprojvc1/az-wus2-v100-32gb-3/b060087b7f6e41d500eec6075e724ae4
% https://ml.azure.com/job/msrhyper/msr-lab-dgx2/53c58ee0586717baaeb06e954539ec2d
Focal-Tiny (Ours) & 39.4/48.8 & 265/291 & \textbf{43.7} & \textbf{65.2} & \textbf{46.7} & \textbf{28.6} & \textbf{47.4} & \textbf{56.9} & \textbf{44.8} & \textbf{67.7} & \textbf{49.2} & \textbf{41.0} & \textbf{64.7} & \textbf{44.2} \\
% https://ml.azure.com/job/msrhyper/msr-lab-hyper-dgx2/4ef4f48228c8438937d71a765ba93af8
% https://ml.azure.com/job/msrhyper/msr-lab-dgx2/3decd79753d86cf3169aa284a6090bf9
% Focal-T/14 & - & - &  \\
\midrule
ResNet101~\cite{he2016deep} & 56.7/63.2 & 315/336 & 38.5 & 57.8 & 41.2 & 21.4 & 42.6 & 51.1 & 40.4 & 61.1 & 44.2 & 36.4 & 57.7 & 38.8 \\
ResNeXt101-32x4d~\cite{xie2017aggregated} & 56.4/62.8 & 319/340 & 39.9 & 59.6 & 42.7 & 22.3 & 44.2 & 52.5 & 41.9 & 62.5 & 45.9 & 37.5 & 59.4 & 40.2 \\
PVT-Medium~\cite{wang2021pyramid} & 53.9/63.9 & 283/302 & 41.9 & 63.1 & 44.3 & 25.0 & 44.9 & 57.6 & 42.0 & 64.4 & 45.6 & 39.0 & 61.6 & 42.1 \\
ViL-Medium~\cite{zhang2021multi} & 50.8/60.1 & 339/261 & 42.9 & 64.0 & 45.4 & 27.0 & 46.1 & 57.2 & 43.4 & 65.9 & 47.0 & 39.7 & 62.8 & 42.1 \\
Swin-Small~\cite{liu2021swin} & 59.8/69.1 & 335/354 & 45.0 & 66.2 & 48.3 & 27.9 & 48.8 & 59.5 & 46.5 & 68.7 & 51.3 & 42.1 & 65.8 & 45.2 \\
% https://ml.azure.com/job/msrhyper/msr-lab-dgx2/867dfc757f15c3dd13515fba83b096a5
% https://ml.azure.com/job/msrhyper/msr-lab-dgx2/f3094efd21955f989585a55c63a6f3a1
Focal-Small (Ours) & 61.7/71.2 & 367/401 & \textbf{45.6} & \textbf{67.0} & \textbf{48.7} & \textbf{29.5} & \textbf{49.5} & \textbf{60.3} & \textbf{47.4} & \textbf{69.8} & \textbf{51.9} & \textbf{42.8} & \textbf{66.6} & \textbf{46.1} \\

% 47.2634,67.7992,51.0353,31.5859,50.8644,61.1443
% https://ml.azure.com/runs/swin_Transformer-33e2a88e-well-ray-trainval_retina_with_swin_small-33714287?wsid=/subscriptions/46da6261-2167-4e71-8b0d-f4a45215ce61/resourcegroups/msrhyperVC/workspaces/msrhyper&tid=72f988bf-86f1-41af-91ab-2d7cd011db47#outputsAndLogs
% Focal-S/14 & - & - &  \\
\midrule
ResNeXt101-64x4d~\cite{xie2017aggregated} & 95.5/102 & 473/493 & 41.0 & 60.9 & 44.0 & 23.9 & 45.2 & 54.0 & 42.8 & 63.8 & 47.3 & 38.4 & 60.6 & 41.3 \\
PVT-Large\cite{wang2021pyramid} & 71.1/81.0 & 345/364 & 42.6 & 63.7 & 45.4 & 25.8 & 46.0 & 58.4 & 42.9 & 65.0 & 46.6 & 39.5 & 61.9 & 42.5 \\
ViL-Base~\cite{zhang2021multi} & 66.7/76.1 & 443/365 & 44.3 & 65.5 & 47.1 & 28.9 & 47.9 & 58.3 & 45.1 & 67.2 & 49.3 & 41.0 & 64.3 & 44.2 \\
Swin-Base~\cite{liu2021swin} & 98.4/107 & 477/496 & 45.0 & 66.4 & 48.3 & 28.4 & 49.1 & 60.6 & 46.9 & 69.2 & 51.6 & 42.3 & 66.0 & 45.5 \\
% https://ml.azure.com/job/msrhyper/msr-lab-dgx2/24d03a24deb92bae34180861e79be137
% https://ml.azure.com/job/msrhyper/msr-lab-dgx2/2dbd6318c7572bab2b305c1eb8c0e08e
Focal-Base (Ours) & 100.8/110.0 & 514/533 & \textbf{46.3} & \textbf{68.0} & \textbf{49.8} & \textbf{31.7} & \textbf{50.4} & \textbf{60.8} & \textbf{47.8} & \textbf{70.2} & \textbf{52.5} & \textbf{43.2} & \textbf{67.3} & \textbf{46.5} \\
% 46.9125,67.8018,50.2914,31.8636,50.3402,61.4545
% https://ml.azure.com/job/msrhyper/msr-lab-dgx2/738a9291c2527cd3faad9233b3577d6b
% Focal-B/14 & - & - &  \\
\bottomrule
\end{tabular}}
\end{center}
\vspace{1pt}
\caption{COCO object detection and segmentation results with RetinaNet~\cite{lin2017focal} and Mask R-CNN~\cite{he2016deep} trained with 1x schedule. This is a full version of Table~\ref{tab:object_detection}. The numbers before and after ``/'' at column 2 and 3 are the model size and complexity for RetinaNet and Mask R-CNN, respectively.}
\label{tab:object_detection_1x}
\vspace{-2mm}
\end{table}
\vspace{-5pt}

\subsection{Model inspections}
\begin{figure}[t]
\begin{minipage}{0.48\linewidth}
\centering
\footnotesize
  \includegraphics[width=1.0\textwidth]{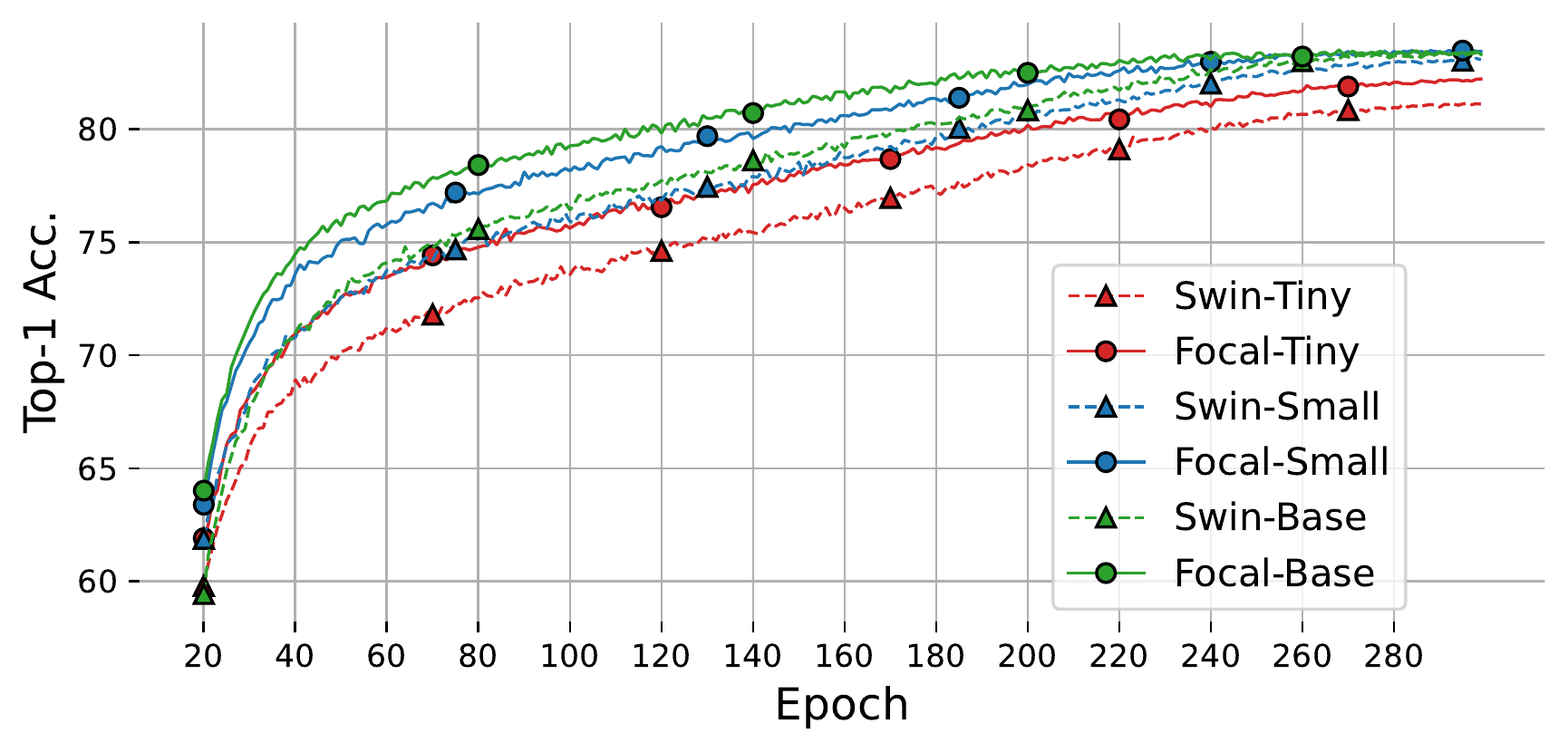}
  \captionof{figure}{Training curves (Top-1 validation Acc.) for image classification with Swin Transformers and our Focal Transformers.}
  \label{fig:cls_acc_with_epochs}
  \vspace{-3pt}
\end{minipage}
\hfill
\begin{minipage}{0.48\linewidth}
\centering
\footnotesize
  \includegraphics[width=1.0\textwidth]{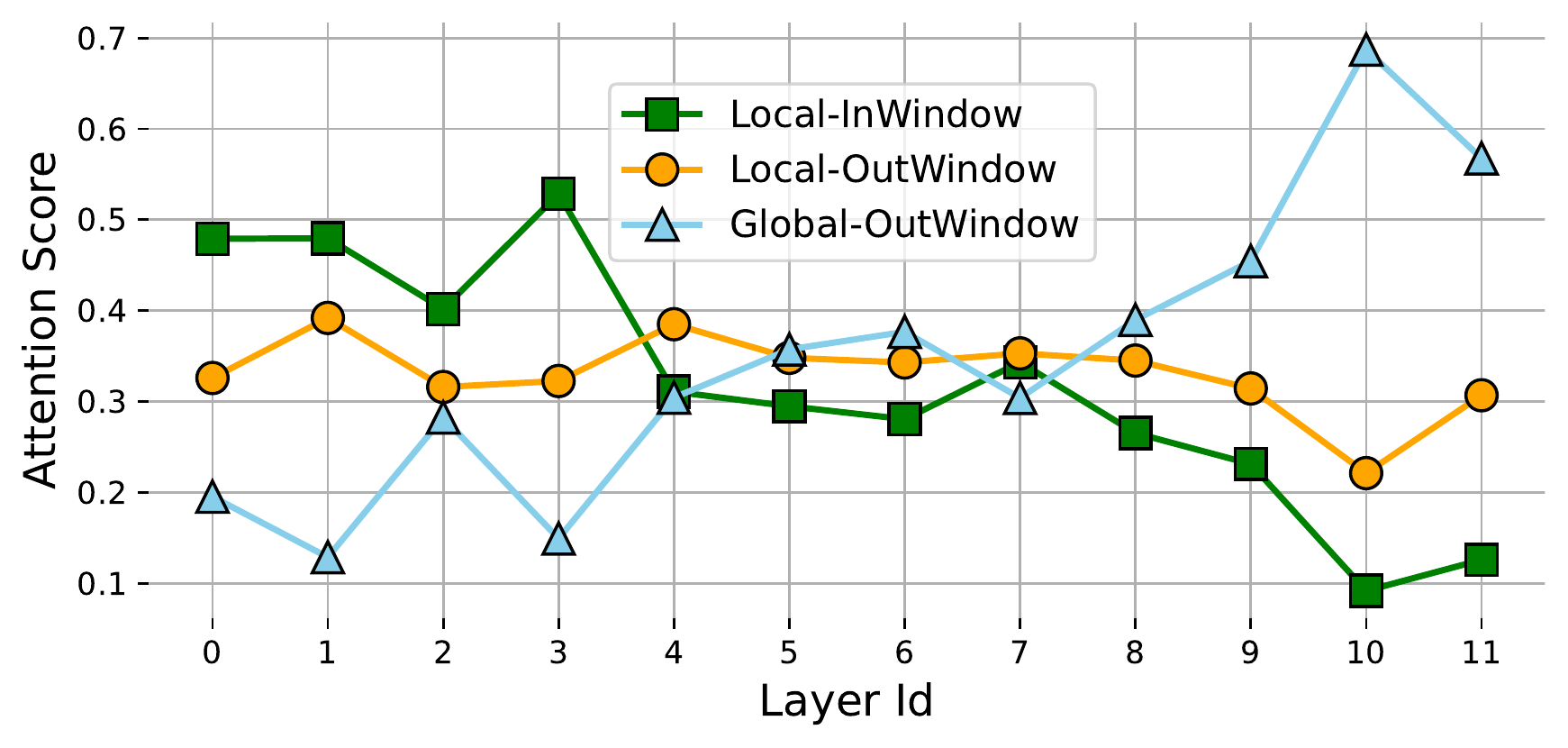}
  \captionof{figure}{Summed up attention score at each layer for: a) local tokens inside window; b) local tokens surrounding window and c) global tokens.}
  \vspace{-3pt}
  \label{fig:local_global_attention_score}
\end{minipage}
% \hfill
% \begin{minipage}{0.5\linewidth}
% \setlength{\tabcolsep}{1.8pt}
% \centering
% \footnotesize
% \resizebox{0.96\linewidth}{!}{
%   \begin{tabular}{cc|ccccc}
%     \toprule
%     Depths & Model & \#Params. &  FLOPs & Top-1~(\%) & $AP^b$ & $AP^m$ \\
%     \midrule
%     \multirow{2}{*}{2-2-2-2}   & Swin  & 21.2 & 3.1 & 78.7 & 38.2 & 35.7  \\
%       & Focal  & 21.7 & 3.4 & 79.9 & 40.5 & 37.6  \\
%       \midrule
    
%     \multirow{2}{*}{2-2-4-2}                             & Swin & 24.7 & 3.8 & 80.2 & 41.2 & 38.1  \\
%                                  & Focal & 25.4 & 4.1 & \cellcolor{blue!25}{81.4} & 43.3 & \cellcolor{blue!25}{39.8}  \\    
%       \midrule
%     \multirow{2}{*}{2-2-6-2}                             & Swin & 28.3 & 4.5 & \cellcolor{blue!25}{81.2} & 43.7 & \cellcolor{blue!25}{39.8}  \\                      
%                                  & Focal & 29.1 & 4.9 & 82.2 & 44.8 & 41.0  \\                                 
%   \end{tabular}}
%   \vspace{2pt}
%   \captionof{table}{Model performance with the change of model depth.}
%   \label{tab:performance_with_depth}
% \end{minipage}
% \vspace{-5pt}
\end{figure}

\textbf{Learning speed comparison}. As we briefly discussed earlier, our model shows faster learning speed on object detection task. In Fig.~\ref{fig:cls_acc_with_epochs}, we show the top-1 validation accuracy of our models and Swin Transformers for image classification task. Accordingly, our Focal Transformers have much faster learning speed as well. For example, Focal-Tiny has 75.7\% top-1 accuracy at 100-th epoch while Swin-Tiny has 73.9\% top-1 accuracy. Similarly, Focal-Small achieves 78.3\% at 100-th epoch, which is 2.0 point higher than Swin-Small. Even for the base models, this gap is still maintained for a long duration until the end of the training. We attribute this faster learning speed to the long-range interactions introduce by our focal attention mechanism in that it can help to capture the global information at very beginning.

\textbf{Attention scores for different token types}. In our main submission, we have shown both local and global attentions are important. Here, we study how much local and global interactions occur at each layer. Using Focal-Tiny trained on ImageNet-1K as the target, we show in Fig.~\ref{fig:local_global_attention_score} the summed up attention scores for three type of tokens: 1) local tokens inside the window; 2) local tokens surrounding the window and 3) global tokens after the window pooling. To compute these scores, we average over the all local windows and then also take the average over all heads. Finally, we sum up the attention scores that belongs to the aforementioned three type of tokens. These attention scores are further averaged over the whole ImageNet-1K validation set. In Fig.~\ref{fig:local_global_attention_score}, we can see a clear trend that the global attention becomes stronger when it goes to upper layers, while the local attention inside a window is weakened gradually. This indicates that: 1) our model heavily relies on both short- and long-range interactions. Neither of them are neglected in the model at all layers and stages; 2) the gradually strengthened global and weakened local attentions indicate that model tends to focus on more local details at earlier stages while on more global context at the later stages.

% \begin{figure}[t]
%   \includegraphics[width=.33\textwidth]{figures/attention_local2global_s1_h1.pdf}
%   \includegraphics[width=.33\textwidth]{figures/attention_local2global_s1_h2.pdf}
%   \includegraphics[width=.33\textwidth]{figures/attention_local2global_s1_h3.pdf}
%   \caption{Learned relative position bias between local window and the global tokens in Focal-Tiny trained on ImageNet-1K. We show the three heads at the first layer of first stage. They are $7\times 7$ bias matrices since the focal region size is set to 7.}
%   \label{fig:local2global_rpb_s1}
% \end{figure}

\begin{figure}[t]
\begin{subfigure}{\textwidth}
  \includegraphics[width=0.5\textwidth]{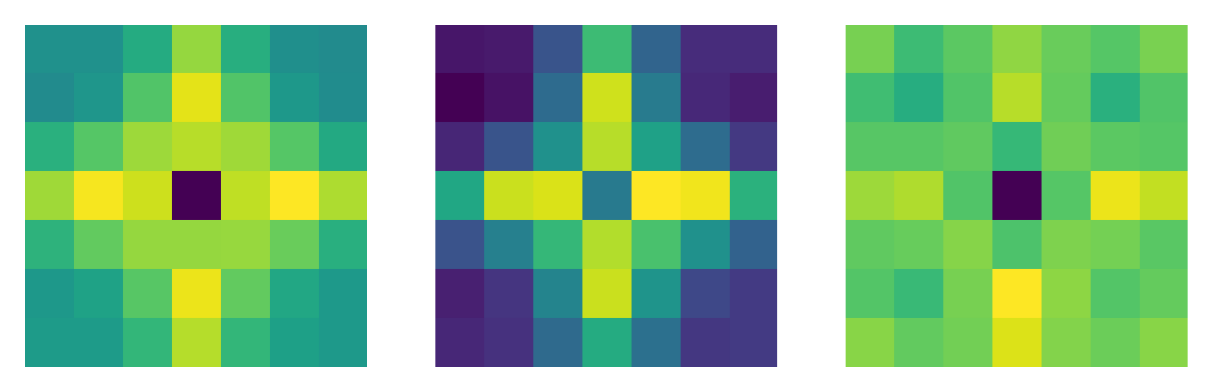}
  \includegraphics[width=0.5\textwidth]{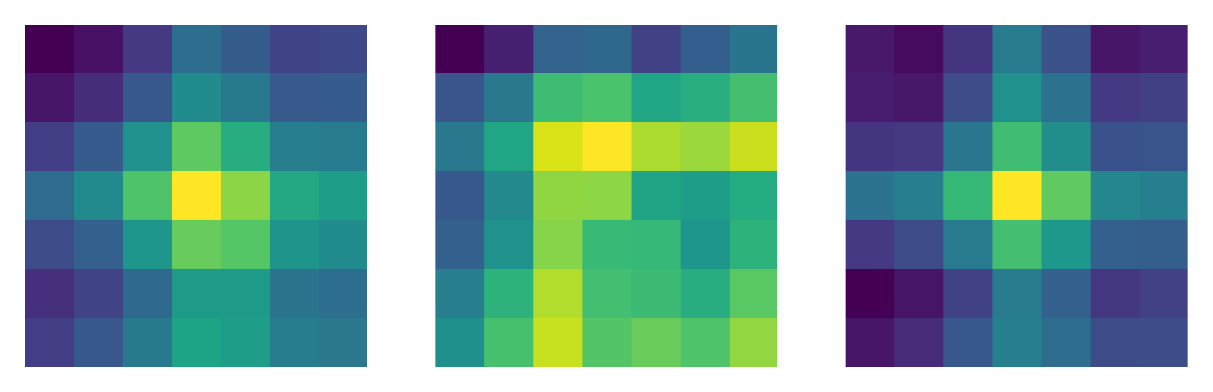}  
  \caption{Stage 1, left 3 for first layer, right 3 for second layer, size=$7\times 7$}
\end{subfigure}
\begin{subfigure}{\textwidth}
  \includegraphics[width=\textwidth]{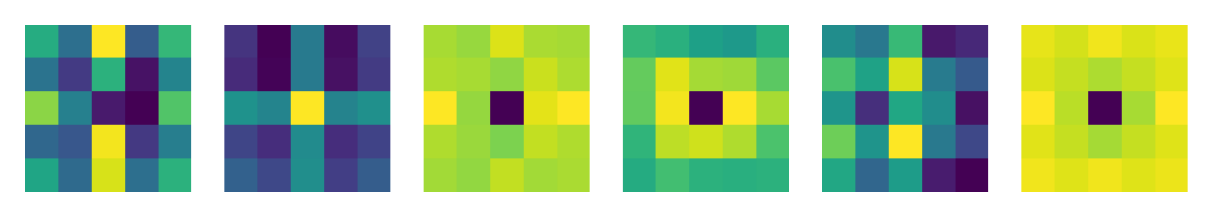}
  \includegraphics[width=\textwidth]{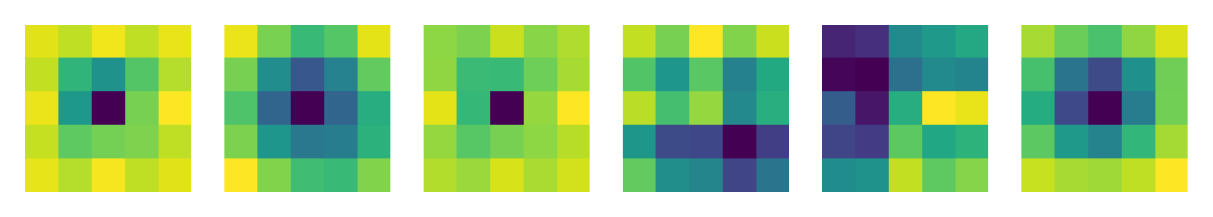}  
\caption{Stage 2, top row for first layer and bottom row for second layer, 6 heads, size=$5\times 5$}
\end{subfigure}
\begin{subfigure}{\textwidth}
  \includegraphics[width=\textwidth]{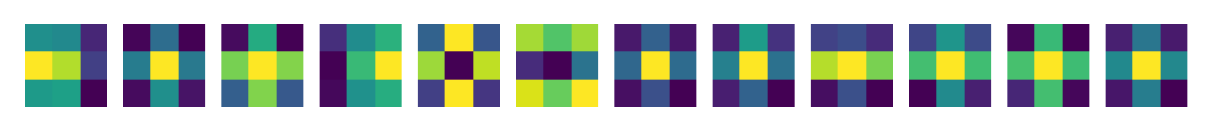}
  \includegraphics[width=\textwidth]{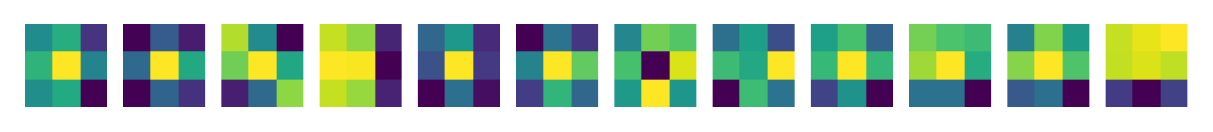}    
  \includegraphics[width=\textwidth]{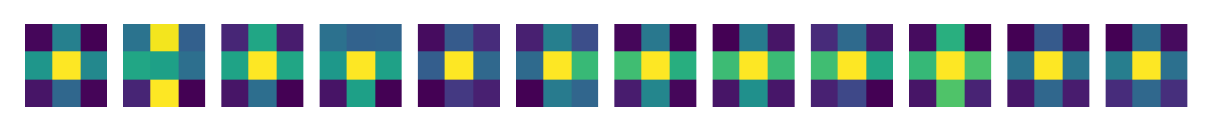}    
  \includegraphics[width=\textwidth]{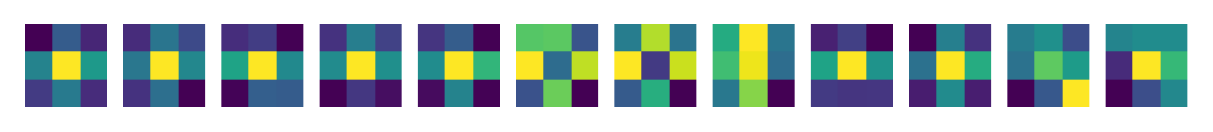}    
  \includegraphics[width=\textwidth]{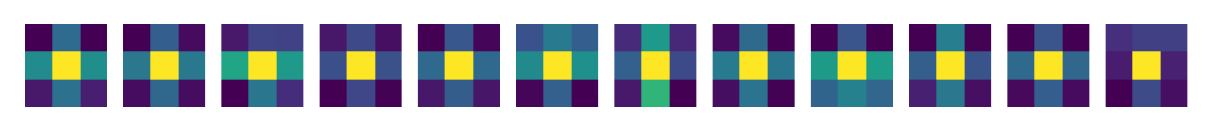}    
  \includegraphics[width=\textwidth]{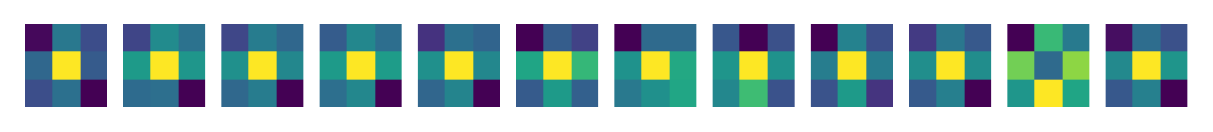}      
\caption{Stage 3, 6 layers from top to bottom row, 12 heads, size=$3\times 3$}
\end{subfigure}
  \caption{Learned relative position bias between local window and the global tokens in Focal-Tiny trained on ImageNet-1K. From top to bottom, we show the learned relative position bias for all heads at (a) stage 1, (b) stage 2 and (c) stage 3. Since the focal region size is 1 for stage 4 in classification models, we only show the first three stages.}
  \label{fig:local2global_rpb_imagenet}
\end{figure}

\begin{figure}[t]
\begin{subfigure}{\textwidth}
  \includegraphics[width=0.5\textwidth]{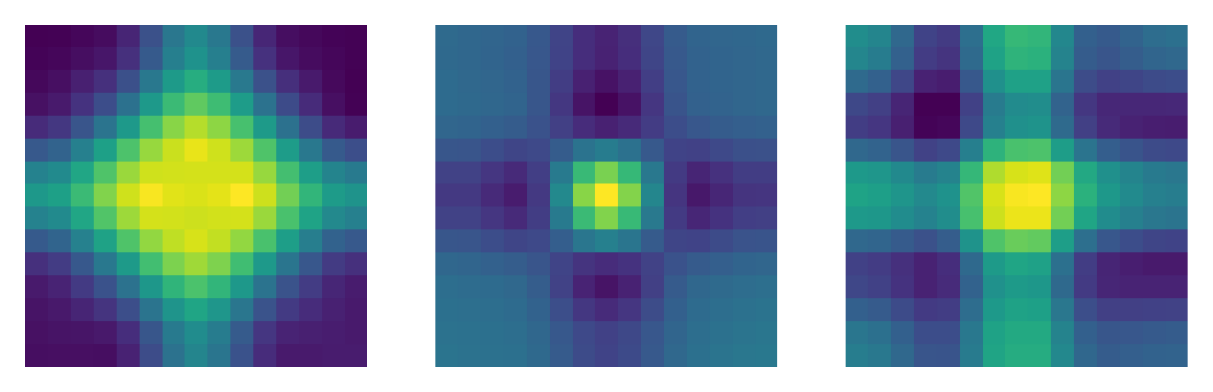}
  \includegraphics[width=0.5\textwidth]{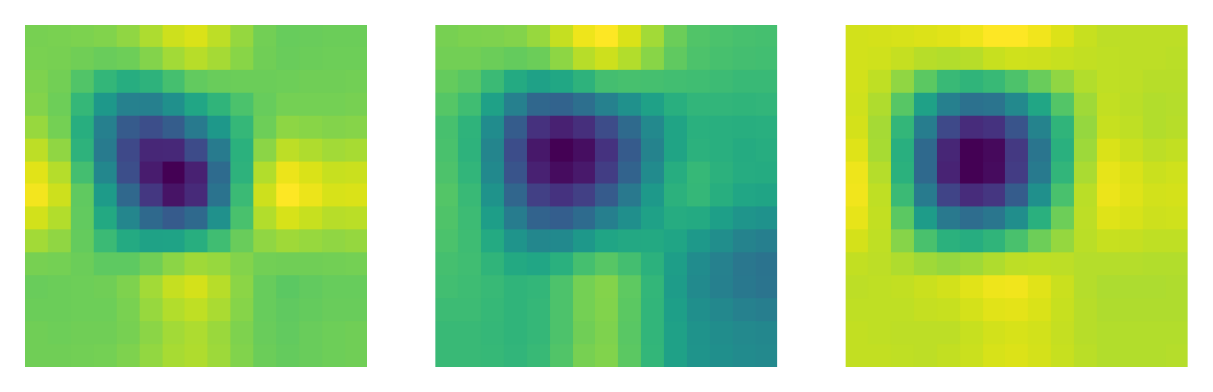}  
  \caption{Stage 1, left 3 for first layer and right 3 for second layer, 3 heads, size=$15\times 15$}
\end{subfigure}
\begin{subfigure}{\textwidth}
  \includegraphics[width=\textwidth]{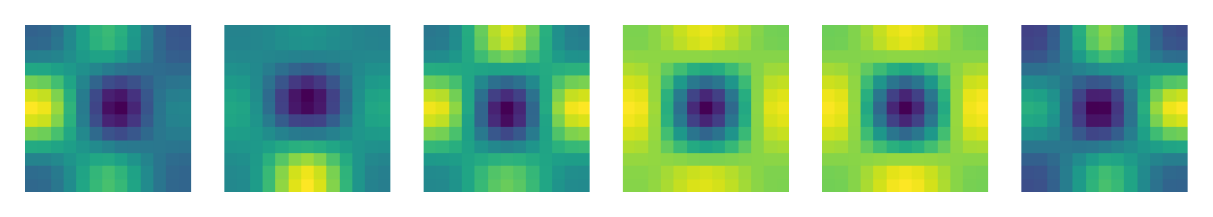}
  \includegraphics[width=\textwidth]{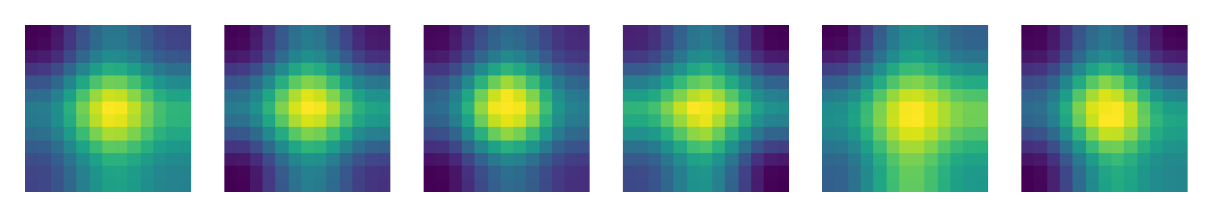}  
\caption{Stage 2, top row for first layer and bottom row for second layer, 6 heads, size=$13\times 13$}
\end{subfigure}
\begin{subfigure}{\textwidth}
  \includegraphics[width=\textwidth]{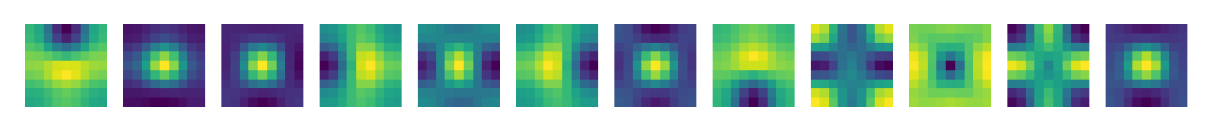}
  \includegraphics[width=\textwidth]{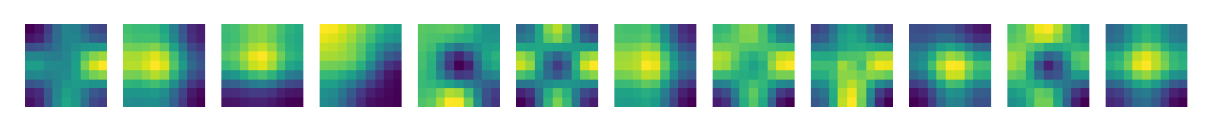}    
  \includegraphics[width=\textwidth]{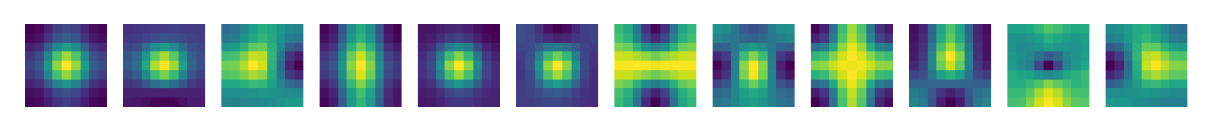}    
  \includegraphics[width=\textwidth]{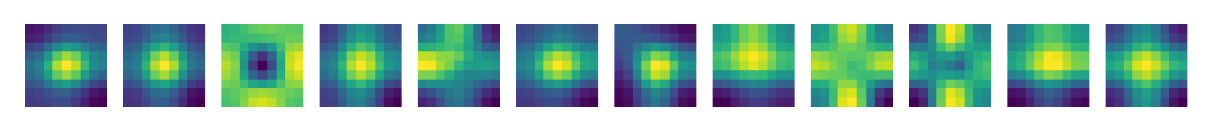}    
  \includegraphics[width=\textwidth]{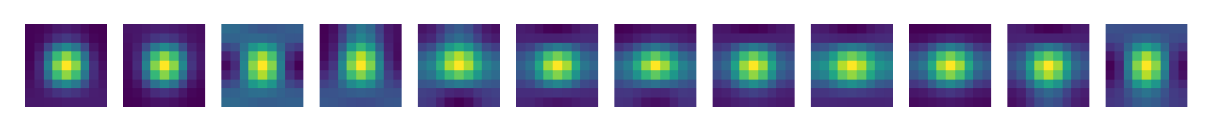}    
  \includegraphics[width=\textwidth]{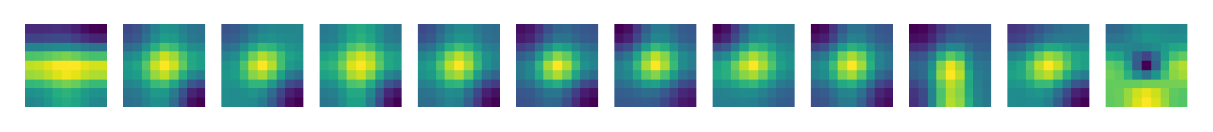}      
\caption{Stage 3, 6 layers from top to bottom row, 12 heads, size=$9\times 9$}
\end{subfigure}
\begin{subfigure}{\textwidth}
  \includegraphics[width=\textwidth]{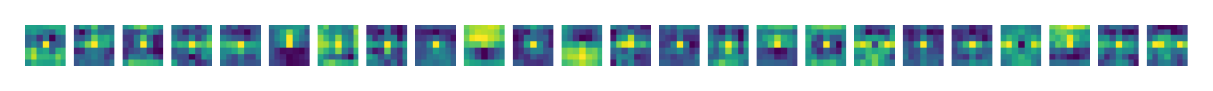}    
  \includegraphics[width=\textwidth]{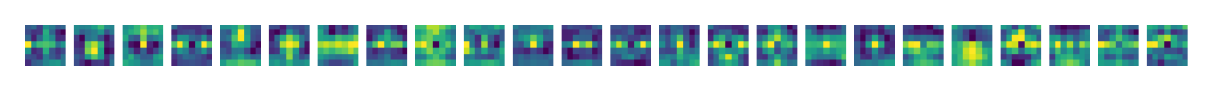}      
  \caption{Stage 4, top row for first layer and bottom row for second layer, 24 heads, size=$7\times 7$}
\end{subfigure}
  \caption{Learned relative position bias between local window and the global tokens in Focal-Tiny for object detection trained on COCO. From top to bottom, we show the relative position bias for different heads at (a) stage 1, (b) stage 2, (c) stage 3 and (d) stage 4.}
  \label{fig:local2global_rpb_coco}
\end{figure}

% \textbf{Window Pooling Weights}.

% show our focal transformer has a faster convergence on both image classification and object detection. show the curves.

\textbf{Local-to-global relative position bias}. We further inspect what our model learns for the local to global relative position bias introduced in Eq.~\eqref{eq:fsa}. This relative position bias is a good indicator on how the model put its attention weight on local and global regions. In our Focal Transformers, the focal region sizes at four stages are $(7,5,3,1)$ and $(15,13,9,7)$ for image classification and object detection, respectively. In Fig.~\ref{fig:local2global_rpb_imagenet} and Fig.~\ref{fig:local2global_rpb_coco}, we visualize the learned relative position bias matrices for all heads and all layers in our Focal-Tiny model trained on ImageNet-1K and COCO, respectively. Surprisingly, though all are randomly initialized, these relative position biases exhibit some interesting patterns. At the first stage of image classification model, all three heads learn to put much less attention on the center window at first layer while focus more on the center at the second layer. For object detection model, however, they are swapped so that the first layer focus more on the center part while the second layer learns to extract the global context from surrounding. As a result, these the two layers cooperate with each other to extract both local and global information. At the second stage of both models, we observe similar property that the two consecutive layers have both local and global interactions. Compared with image classification model, the object detection model has more focus on the center regions. We suspect this is because object detection needs to extract more fine-grained information at local regions to predict the object category and location. At the third stage, we can see there is a fully mixture of local and global attentions in both models. Surprisingly, though randomly initialized, some of the heads automatically learn to disregard the center window pooled token which has much redundancy with the fine-grained tokens inside the center window.

\end{document}